%% file: robustregret_arxiv.tex
\documentclass[11pt]{article}

\usepackage{macros}

\newlength\titlebox 
\usepackage[symbol]{footmisc}

\begin{document}
\title{On Misspecification in Prediction Problems \\
  and Robustness via Improper Learning}
\author{Annie Marsden \footnote{Supported by ACM SIGHPC/Intel Fellowship} \and John Duchi \footnote{Supported by NSF CAREER Award CCF-1553086, ONR YIP N00014-19-2288, and the DAWN Consortium} \and Gregory Valiant \footnote{Supported by DOE award DE-SC0019205, NSF awards 1813049 and 1704417, and ONR YIP N00014-18-1-2295}}
\date{Stanford University}
\maketitle

\input{sections/abstract}

\section{Introduction}
\subfile{sections/introduction}

\input{sections/relatedwork}

\input{sections/lowerbound}

\input{sections/upperbound}

\input{sections/experiments}

\input{sections/discussion}

\bibliography{bib}
\bibliographystyle{abbrvnat}

\appendix

\onecolumn

\subfile{sections/proof-lowerbound}

\subfile{sections/appendix-linearlemma}
\subfile{sections/proof-perturbation-separation}

\subfile{sections/proof-slow-rates-oops}

\section{Technical appendices}


\subfile{sections/appendix-mixconstproofs}

\subfile{sections/appendix-optseqglm2}

\subfile{sections/appendix-bvm}

\subfile{sections/appendix-bvmcor}

\end{document}

%% file: sections/abstract.tex
\begin{abstract}
  We study probabilistic prediction games when the underlying model is
  misspecified, investigating the consequences of predicting using an
  incorrect parametric model. We show that for a broad class of loss
  functions and parametric families of distributions, the regret of playing
  a ``proper'' predictor---one from the putative model class---relative to
  the best predictor in the same model class has lower bound scaling at
  least as
  $\sqrt{\perturbsize n}$, where $\perturbsize$ is a measure of the model
  misspecification to the true distribution in terms of total variation
  distance. In contrast, using an aggregation-based (improper) learner, one
  can obtain regret $d \log n$ for any underlying generating distribution,
  where $d$ is the dimension of the parameter; we exhibit instances in which
  this is unimprovable even over the family of all learners that may play
  distributions in the convex hull of the parametric family.  These results
  suggest that simple strategies for aggregating multiple learners together
  should be more robust, and several experiments conform to this hypothesis.
\end{abstract}

%% file: sections/introduction.tex

Suppose we seek a probability distribution $p(y \mid x)$ modeling outcomes
$y$ given data $x$. The typical approach is to choose a parametric family of
probability distributions, then find the ``best'' member of this family
according to a given loss.  It is rarely realistic to assume that the
parametric family is well-specified, and thus it is important to understand
the consequences of misspecification and how to circumvent these downsides.
To address these challenges, in this paper we derive a new measure of a
problem's robustness to misspecification that relies on the curvature of
the loss at hand and putative parametric family, proving that this measure
lower bounds convergence rates for prediction error and certifies the
\emph{failure} of a parametric family and loss to be robust (or achieve
optimal convergence rates for prediction).  To complement this new family of
lower bounds for probabilistic prediction problems, we build out of earlier
work on \emph{improper learning}~\citep{Vovk98, FosterKaLuMoSr18}---when we
may choose predictions $p(y \mid x)$ outside the given model family---to
show how it is possible to be robust to such misspecification, and
moreover, we give new optimality guarantees for such improper procedures.



Formalizing our setting, we consider the following probabilistic
game: a player receives a covariate vector $x \in \mc{X}$, plays a
distribution $p(\cdot \mid x)$ on a target set $\mc{Y}$, then receives $y
\in \mc{Y}$ and suffers loss
\begin{equation*}
  \loss(p(\cdot \mid x), y).
\end{equation*}
We study both a sequential and a stochastic variant of this
problem. In the former, for a sequence of examples $\{(x_i,
y_i)\}_{i=1}^n$, a player chooses a distribution $p_k$ depending on the past
examples $\{(x_i, y_i)\}_{i = 1}^{k-1}$, and then for a fixed conditional
distribution $p$ on $Y \mid X$, suffers regret
\begin{equation*}
  \regret(p) \defeq \sum_{i = 1}^n \loss(p_i(\cdot \mid x_i), y_i)
  - \sum_{i = 1}^n \loss(p(\cdot \mid x_i), y_i).
\end{equation*}
In the stochastic variant,
the examples $(x_i, y_i)$ are i.i.d.\ from an unknown distribution $P$, and
we consider the risk of the conditional p.m.f.\ $p$,
\begin{equation*}
  \risk_P(p) \defeq \E[\loss(p(Y \mid X))]
  = \int \loss(p(\cdot \mid x), y) dP(x, y),
\end{equation*}
where the expectation is taken over $(X, Y) \sim P$.  The goal is to play
$p_i$ or $p$ above to make the regret
and risk as small as possible.

This regret and risk minimization framework is familiar from the universal
prediction and probabilistic forecasting literature~\citep{MerhavFe98,
  Grunwald07, GneitingRa07, CoverTh06, CesaBianchiLu06}, which considers
best possible estimators and online learners for the regret $\regret$ over
various losses $\loss$ and families $\mc{P}$ of possible predictive
distributions.  In this paper, we study these regret and risk minimization
formulations over parametric families of distributions $\{p_\theta(\cdot
\mid x)\}_{\theta \in \Theta}$, where $\Theta \subset \R^d$ is a convex set.
We shall either consider the regret
\begin{subequations}
  \begin{equation}
    \label{eqn:regret}
    \regret^\Theta \defeq \sup_{\theta \opt \in \Theta}
    \regret(p_{\theta\opt})
    = \sum_{i = 1}^n \loss(p_i(\cdot \mid x_i), y_i)
    - \inf_{\theta\opt \in \Theta} \sum_{i = 1}^n \loss(p_i(\cdot \mid x_i), y_i)
  \end{equation}
  or---in the stochastic version of the problem---the excess risk relative
  to this family,
  \begin{equation}
    \label{eqn:excess-risk}
    \risknorm_P(p) \defeq \risk_P(p) - \inf_{\theta\opt \in \Theta} \risk_P(p_{\theta\opt}).
  \end{equation}
\end{subequations}
When the conditional distribution of $Y \mid X$ belongs to the parametric
family $\param$ where $\Theta \subset \R^d$, maximum likelihood estimators
enjoy rates of convergence of $O(d/n)$ for the excess
risk~\eqref{eqn:excess-risk} as $n$ grows~\cite{VanDerVaart98}. In typical
practice, however, the data generating distribution is misspecified, so it
is important to understand how this misspecification impacts possible
convergence rates and optimal estimators.

We thus consider three intertwined objects: the parametric family $\param$
against which we compare the performance of our prediction $p$, a family
$\playdists$ of distributions on $Y$ given $X$ that we may play
(i.e.\ predict from), and the family $\naturedists$ of data generating
distributions that nature may choose.  We study the interaction between
these three and the impact of allowing the family $\naturedists$
to differ from the parametric model $\param$.
The traditional approach considers the minimax
excess risk over the family $\Theta$,
\begin{equation}
  \label{eqn:traditional-minimax}
  \inf_{\what{p}_n} \sup_{\theta \in \Theta}
  \E_{P_\theta^n}\left[\risknorm_P(\what{p}_n)\right],
\end{equation}
where the infimum is over all estimators $\what{p}_n$ that use the $n$
points $\{(X_i, Y_i)\}_{i=1}^n$ to output a distribution $p(Y \mid X)$, and
the expectation is taken over $\{(X_i, Y_i)\}_{i=1}^n \simiid P_\theta$,
where we have abused notation to use $P_\theta$ to denote the joint over
$(X, Y)$ when $Y \mid X = x$ follows $p_\theta(\cdot \mid x)$.  We elaborate
this setting slightly. First, we restrict the estimator $\what{p}_n$ to take
values in a set $\Gamma$ of distributions (for example, we might take
$\Gamma = \pt$, the parametric family, or its
convex hull), which we write as $\what{p}_n \in \Gamma$. Second, we
expand the supremum~\eqref{eqn:traditional-minimax} to also include
distributions $P$ \emph{near} the model $P_\theta$: recalling the definition
of the total-variation distance $\tvnorm{P- Q} \defeq \sup_A |P(A) - Q(A)|$,
we consider distributions $P$ for which there is some $\theta \in \Theta$
such that $\tvnorm{P - P_\theta} \le \perturbsize$. This gives us our
misspecified minimax risk.
\begin{definition}
  \label{def:minimax-risk}
  Let $\Theta \subset \R^d$, $\perturbsize \ge 0$, and
  $\playdists$ be a set of allowable distributions $p(Y \mid X)$.
  The \emph{minimax risk} at variation distance $\perturbsize$ is
  \begin{equation}
    \label{eqn:minimax-risk}
	 \minimax_n(\Theta, \playdists, \perturbsize)
      \defeq  \inf_{\what{p}_n \in \Gamma}
    \sup_{\theta \in \Theta}
    \sup_{P : \tvnorms{P - P_\theta} \le \perturbsize} 
    \E_{P^n}[\risknorm_P(\what{p}_n)].
  \end{equation}
\end{definition}
\noindent
The quantity~\eqref{eqn:minimax-risk} is somewhat complex. The
idea is to quantify---via the parameter $\perturbsize$---the impact of
allowing the data generating distribution $P$ to depart slightly from
the parametric family $\param$ while constraining
ourselves to play a prediction from the family $\playdists$.

The typical setting in online convex optimization and
learning~\citep{CesaBianchiLu06} is to take the family of ``playable''
distributions to be the parametric family $\playdists = \param$.  In this
case, standard minimax risk bounds show that in the well-specified setting
that the data comes from the parametric family (i.e.\ $\perturbsize = 0$ in
Def.~\ref{def:minimax-risk}) and the loss $\loss$ is smooth, then we expect
the risk to scale as $d / n$ (cf.~\cite{VanDerVaart98, Wainwright19,
  Bach14}).  Yet as we show in the first part of this work, such results
need not be stable to perturbations away from the parametric model.  We show
that the curvature of the loss relative to predictions and the parameter
space $\Theta$ essentially governs convergence rates: when losses are
appropriately ``flat,'' there is little information and rates are
necessarilty slow and misspecification carries a potentially heavy penalty;
conversely, when there is substantial curvature, rates exhibit
less antagonistic behavior. Accordingly, we introduce
what we term the \emph{linearity constant} $\textup{Lin}$ of the loss $\loss$,
family $\pt$, and misspecification $\perturbsize$, showing
a lower bound of roughly $\min\{1/\sqrt{n}, \textup{Lin} / n\}$
on the minimax risk~\eqref{eqn:minimax-risk}. In some cases we delineate,
$\textup{Lin}$ may be exponentially large in problem parameters,
so convergence rates slow to the worst-case rates for general
online convex optimization~\citep{Zinkevich03,
  ShalevShSrSr09, AgarwalBaRaWa12}, and we consider the family
sensitive to misspecification.

To complement these negative results, we highlight a solution to this
instability by considering the convex hull of the parameteric family, that
is the set of mixtures, aggregations, or ensembles $\conv \param \defeq \{
\int_{\theta \in \Theta} p_{\theta} d \mu (\theta) \tr{ s.t. } \int_{\theta
  \in \Theta} d \mu(\theta) =1 \tr{ and } d\mu \ge 0\}$.  The idea to
combine probabilistic forecasts is classical~\citep{GrangerRa84, GenestZi86,
  Jacobs95, ClemenWi99, HallMi07}. When the loss function is \emph{mixable}
(which we define later), Vovk's Aggregating Algorithm and its variants,
e.g.\ exponential weights, $\mathsf{Exp}3$, and Bayesian universal
prediction~\citep{Vovk90, Vovk92, Vovk98, Grunwald07, CesaBianchiLu06,
  AuerCeFrSc02}, provide stability and achieve minimax regret $O( d \log n)$
for any $\perturbsize$ in Definition~\ref{def:minimax-risk}. By a standard
online-to-batch conversion (Jensen's inequality)~\cite{CesaBianchiCoGe02},
this guarantees a minimax excess risk~\eqref{eqn:minimax-risk}
of at most $O (d \log n /n)$. We also
show that for generalized linear models, Vovk's Aggregating Algorithm is
optimal up to log-factors with respect to the misspecification parameter
$\perturbsize$. That is, there is no better algorithm to use if you are
guaranteed certain values of $\perturbsize$.

It is perhaps unfair to consider the entire convex hull of $\param$ for the
class $\playdists$, as this could potentially yield much smaller risk than
the parametric family in the risk~\eqref{eqn:excess-risk}. Indeed, we give
an example in which the best parameteric predictor has no predictive power,
while returning a mixture of two parameterized distributions achieves zero
loss (though we also give examples in parametric families where considering
the convex hull provides no benefit).  To justify the aggregation strategy
we give a Bernstein von-Mises theorem under misspecification,
which shows that the strategy converges to a Gaussian centered at the
risk-minimizing parameter estimate $\what{\theta}_n$ with covariance
shrinking at rate $O(1/n)$; a corollary of this is that
Vovk's Aggregating Algorithm returns a distribution which converges in total
variation distance to $p_{\what{\theta}_{n}} \in \{p_\theta\}_{\theta \in
  \Theta}$.
Thus, aggregation (or exponential weights) stabilizes predictions while
asymptotically enjoying identical convergence to standard risk-minimization
procedures.



%% file: sections/relatedwork.tex

\subsection{Related Work}
\label{section:relatedwork}

Our results broadly fall under probabilistic universal prediction in which
the data $(x_{t}, y_{t})$ can be any arbitrary
sequence~\citep{Rissanen84,MerhavFe98, CesaBianchiLu99, CesaBianchiLu06,
  Grunwald07, Cover91}. That Vovk's Aggregating Algorithm provides minimax
rate stability is known~\cite{FosterKaLuMoSr18}, and this is similar to the
minimax guarantees of Bayesian models in universal
prediction~\citep{Grunwald07}; a long line of work gives the same
logarithmic minimax rates~\citep{MerhavFe98, DesantisMaWe88, HausslerBa92,
  Yamanishi95b}. Early work in these prediction problems focuses on the
logarithmic loss $\logloss(p(\cdot), y) = -\log p(y)$, while more recent
work extends these bounds to exp-concave and so-called ``mixable''
losses~\citep{HazanAgKa07, CesaBianchiLu06}. Our results on minimax lower
bounds, distinguishing carefully between well-specified and misspecified
models and proper and improper predictions, are novel.

While our results are general, applying to exponential families and beyond,
related results are available for logistic regression. In this case, for
$\paramrad, \datarad > 0$ we consider $\Theta = \{\theta : \norm{\theta} \le
\paramrad\}$, $\mc{X} \subset \{x : \norm{x} \le \datarad\}$, and let
$\param$ be the family of binary logistic distributions, $p_{\theta}(y \mid
x) = (1+ e^{-y \theta^T x})^{-1}$, with log loss.  \citet{HazanKoLe14} show
that any algorithm returning some $p_{\theta}$ suffers minimax risk (recall
\eqref{eqn:minimax-risk}) $\Omega(\sqrt{\paramrad/n})$ in the regime where
$n = O(\exp{c B})$ for some positive constant $c > 0$, $\datarad = 1$, and
the allowable perturbation $\perturbsize = 1$.  \citet{FosterKaLuMoSr18}
show that Vovk's Aggregating Algorithm \citep{Vovk98} guarantees minimax
risk $ O( d \log(\paramrad n) / n)$, allowing one to sidestep this lower
bound via improper learning, which we also leverage. In the special case of
logistic regression---see Example~\ref{example:logistic} to come in
Section~\ref{sec:log-examples}---a simplification of our results gives lower
bound $\Omega(1) \sqrt{\perturbsize \paramrad \datarad / n}$ if $n \leq
\exp(\datarad \paramrad / 2)$ and $\Omega(1) \exp(2 \paramrad \datarad / 5)
/ n$ otherwise. We thus show that even when the perturbations away from the
parametric family are small, the minimax risk when the set of playable
distributions is $\playdists = \param$ may grow substantially; this
generalizes~\citet{HazanKoLe14}, where $\datarad = 1$ and
$\perturbsize = 1$, and gives somewhat sharper constants.




%% file: sections/lowerbound.tex
\section{Parametric Model Instability}
\label{section:lowerbounds}

Our first step towards understanding sensitivity to misspecification is to
provide optimality guarantees for the minimax risk in
Definition~\ref{def:minimax-risk} when the player can play only elements of
the parametric family of interest, that is, when $\playdists = \param$.
We focus on losses that depend specifically on $\theta^T x$,
where we have
\begin{equation}
  \label{eqn:generalized-linear-loss}
  \loss(p_\theta(\cdot \mid x), y) = \scalarloss(\theta^T x, y)
\end{equation}
for some twice differentiable and convex $\scalarloss : \R \times \mc{Y} \to
\R_+$.  A broad range of models and losses take this form, including all
generalized linear models~\cite{HastieTiFr09}.

\subsection{Main lower bound}

Our key contribution is to lower bound the minimax risk via a quantity we
term the \emph{linearity constant} of the induced loss $\scalarloss$, which
measures the sensitivity of $\scalarloss$ around various points in its
domain. The first component is (roughly) a measure of
the signal contained in $\scalarloss$ for different targets $y$,
where for $t \in \R$, $y \in \mc{Y}$, and $\scalarloss ' (t_0, y_0)$ as shorthand for $\frac{\partial}{\partial t} \scalarloss(t,y) \lvert_{(t,y) = (t_0,y_0)}$ we define
\begin{equation}
  \label{eqn:qworst}
  \qworst(t, y) \defeq
  \sup_{y_0 \in \mc{Y}, \alpha \in [-1,1]}
  \left\{\frac{\alpha \scalarloss'(\alpha t, y_0)}{
    \alpha \scalarloss'(\alpha t, y_0) - \scalarloss'(t, y)}
  ~ \mid ~
  \sign(\alpha \scalarloss'(\alpha t, y_0))
  \neq \sign(\scalarloss'(t, y)) \right\}.
\end{equation}
We always have $\qworst(t, y) \in [0, 1]$. For many cases, this quantity is
a positive numerical constant (e.g.\ for the squared error $\scalarloss(t,
y) = \half (t - y)^2$ with $\mc{Y} = \R$, we have $\qworst(t, y) = 1$).
Then for given radii $\datarad$ and $\paramrad$, misspecification size
$\perturbsize \in [0, 1]$, and sample size $n$, we define
\begin{equation}
  \label{eqn:def-linearconst}
  \linearconstfull
  \defeq \sup_{\stackrel{y \in \mc{Y}}{
      t^2 + \delta^2 \le \frac{\datarad^2 \paramrad^2}{2}}}
  \left\{
  |\scalarloss'(t, y)|
  \min\left\{\delta \sqrt{\perturbsize \qworst(t, y)},
  \frac{|\scalarloss'(t, y)|}{\sup_{|\Delta| \le \delta}
    \scalarloss''(t + \Delta, y)} \frac{\qworst(t, y)}{\sqrt{2n}}
  \right\}
  \right\}.
\end{equation}

This linearity constant roughly measures the extent to which the loss grows
quickly without substantial curvature, that is, $\scalarloss'(t, y)$ is
large while $\scalarloss''(t, y)$ is small.  A heuristic simplification may
help with intuition: by ignoring the $\qworst$ term and perturbation by
$\Delta$ in $\scalarloss''$, we roughly have
\begin{equation}
  \label{eqn:heuristic-display}
  \linearconstfull \stackrel{\textup{heuristically}}{=}
  O(1) \cdot \sup_{y \in \mc{Y}, |t| \le \datarad \paramrad}
  |\scalarloss'(t, y)| \min\left\{ \datarad \paramrad \sqrt{\perturbsize},
  \frac{|\scalarloss'(t, y)|}{\scalarloss''(t, y)} \frac{1}{\sqrt{n}}
  \right\},
\end{equation}
which makes clearer the various relationships.  When the ratio of
$\scalarloss'(t, y)$ to $\scalarloss''(t, y)$ is large, estimation and
optimization are intuitively hard: there is little curvature to help
identify optimal parameters, while small changes in the parameter induce
large changes in the loss (as $\scalarloss'(t, y)$ is large relative to
$\scalarloss''$). The allowable misspecification of the model---via the
parameter $\perturbsize$---means that in the lower bound, an adversary may
essentially put positive mass on those points for which the ratio
$\scalarloss' / \scalarloss''$ is large, so that one must pay this
worst-case cost.

We then have the following theorem, whose proof
we provide in Appendix~\ref{append:lowerbound}.
\begin{theorem}
  \label{thm:lowerbound}
  Let the loss $\loss$ and family $\{p_\theta\}$ satisfy
  Eq.~\eqref{eqn:generalized-linear-loss}, where $\mc{X} = \{x : \norm{x}
  \le \datarad\}$. Consider $\playdists =\param$, where $\Theta = \{\theta :
  \norm{\theta} \le \paramrad\}$. Then
  \begin{equation*}
    \minimax_n(\Theta, \playdists, \perturbsize)
    \geq  \frac{1}{4 \sqrt{n}} \linearconstfull.
  \end{equation*}
\end{theorem}
\noindent
Using the heuristic display~\eqref{eqn:heuristic-display} above can
provide some intuition. When $|\scalarloss'(t, y)| / \scalarloss''(t,y)
\gtrsim \sqrt{n}$, so that the problem has little curvature, the (heuristic)
linearity constant~\eqref{eqn:heuristic-display} scales as $\sup_{t,y}
|\scalarloss'(t, y)| \datarad \paramrad$, which gives the lower bound
$\sup_{t,y} \frac{|\scalarloss'(t, y)| \datarad \paramrad}{\sqrt{n}}$; this
is the familiar worst-case minimax bound for stochastic convex optimization
with Lipschitz objective on a compact domain~\cite{AgarwalBaRaWa12}. As the
worst-case constructions look very little like standard prediction problems,
one might hope (at least in the absence of misspecification) to achieve
better rates; Theorem~\ref{thm:lowerbound} helps to delineate problems
where this may be impossible.



\subsection{Examples with the logarithmic loss}
\label{sec:log-examples}

It is instructive to consider a few examples to build intuition for
Theorem~\ref{thm:lowerbound} beyond the
heuristic~\eqref{eqn:heuristic-display}, as the linearity constant as
defined may be somewhat challenging to work with. As we shall see, however,
its generality allows exploration of many losses,
including various scoring rules~\cite{GneitingRa07}; for this section,
we focus on the common logarithmic loss for four well-known exponential
family models. In what follows we use the following notation: For a set $\Omega$ such that $f,g :\Omega \to \R$ we write $f \gtrsim g$ if there exists a finite numerical constant $C$ such that for any $\omega \in \Omega$, $f(\omega) \geq C g(\omega)$ and we write $f \asymp g$ if $f \gtrsim g \gtrsim f$.

For our first example, we consider logistic regression, showing that
if we must play proper predictions $p_\theta$, parametric $1/n$ rates
are impossible until $n$ is very large or if the radii
$\datarad$ and $\paramrad$ are small.
\begin{example}[Logistic regression]
  \label{example:logistic}
  For logistic regression with logarithmic loss, we have
  $\mc{Y} = \{-1, 1\}$,
  $p_\theta(y \mid x)
  = \frac{1}{1 + \exp(-y \theta^T x)}$, and $\scalarloss(t, y)
  = \log(1 + e^{-ty})$, so that
  \begin{equation*}
    \scalarloss'(t, y) = \frac{-y}{1 + e^{ty}}
    ~~ \mbox{and} ~~
    \scalarloss''(t, y) = \frac{e^{ty}}{(1 + e^{ty})^2}.
  \end{equation*}
  Without loss of generality, let $y = 1$. If $\datarad \paramrad \le 1$,
  then by taking $t = \half \paramrad \datarad $ and $\delta = \half
  \paramrad \datarad$, it is immediate that $\qworst(t, y) \gtrsim 1$, and
  each of $\scalarloss'(t, y)$ and $\scalarloss''(t, y)$ are numerical
  constants. Then we obtain the lower bound
  \begin{equation*}
    \linearconstfull
    \ge c \min\left\{\paramrad \datarad \sqrt{\perturbsize},
    \frac{1}{\sqrt{n}} \right\},
  \end{equation*}
  so that Theorem~\ref{thm:lowerbound} yields minimax lower bound
  $\min\{\frac{\datarad \paramrad \sqrt{\perturbsize}}{\sqrt{n}},
  \frac{1}{n}\}$.

  The more interesting regime is when $\datarad \paramrad \gg 1$---for
  example, in the natural case that the data and parameter radii
  scale with the dimension of the
  problem---so let us assume $\datarad\paramrad \ge 1$.
  Here, take $y = -1$ and $y_0 = 1$, so that
  for any $\alpha \in [0, 1]$ and $t \in \R$ we have
  $\sign(\scalarloss'(t, y)) = 1 \neq -1 = \sign(\scalarloss'(\alpha t, y_0))$.
  Let $\epsilon \in [0, 1]$ to be chosen and set
  $t^2 = (1 - \epsilon) \frac{\datarad^2 \paramrad^2}{2}$ (where $t \ge 0$).
  Then by taking $\alpha = \frac{1}{\datarad \paramrad}$,
  in the definition~\eqref{eqn:qworst} we have
  \begin{align*}
    \qworst(t, y)
    & \ge \frac{\alpha \frac{1}{1 + e^{t \alpha}}}{
      \alpha \frac{1}{1 + e^{t \alpha}} + \frac{1}{1 + e^{-t}}}
    = \frac{1}{1
      + \datarad \paramrad \frac{1 + e^{t \alpha}}{1 + e^{-t}}}
    \ge \frac{1}{1 + (e + 1) \datarad \paramrad}
    \gtrsim \frac{1}{\datarad \paramrad}
  \end{align*}
  and $\scalarloss'(t, y) = \frac{1}{1 + e^{-t}} \ge \half$.
  Thus for all $\delta \in [0, \datarad \paramrad \sqrt{\epsilon/2} ]$,
  the linearity constant has lower bound
  \begin{equation*}
    \linearconstfull \ge c
    \min \left\{\delta \sqrt{\perturbsize  / \datarad \paramrad},
    \frac{1}{\sup_{|\Delta| \le \delta} e^{-t + \Delta}} \frac{1}{\datarad
      \paramrad \sqrt{n}}\right\}
  \end{equation*}
  where $c > 0$ is a numerical constant. Taking
  $\delta = \datarad \paramrad \sqrt{\epsilon/2}$ and $\epsilon
  = 1/9$ then gives
  \begin{equation*}
    \linearconstfull \ge c
    \min\left\{\sqrt{\perturbsize \datarad \paramrad},
    \exp\left( 3 \datarad \paramrad / (5 \sqrt{2})\right)
    \frac{1}{\datarad \paramrad \sqrt{n}} \right\}.
  \end{equation*}
  In particular, if
  $n \le \frac{e^{6 \datarad \paramrad / 5 \sqrt{2}}}{\datarad^2 \paramrad^2}$,
  then
  $\linearconstfull \ge c \sqrt{\perturbsize \datarad \paramrad}$,
  and otherwise (as $e^x / x \gtrsim e^{.99 x}$ for all $x \ge 1$)
  $\linearconstfull \ge \exp(\datarad \paramrad / 4) / \sqrt{n}$,
  giving us the minimax lower bound
  \begin{equation}
    \label{eqn:logistic-lower-bound}
    \minimax_n(\Theta, \playdists, \perturbsize)
    \ge c \min\left\{\frac{\sqrt{\perturbsize \datarad \paramrad}}{
      \sqrt{n}},
    \frac{\exp(2 \datarad \paramrad / 5)}{n} \right\}.
  \end{equation}

  We may contrast this lower bound with previous results.  In the regime
  where $\perturbsize = 1$, \citet{HazanKoLe14} show that for $\datarad = 1$
  and numerical constants $c_0, c_1 > 0$, any algorithm playing
  parametric predictors $p_\theta$ necessarily suffers minimax risk
  $\Omega(\sqrt{\paramrad / n})$ whenever $n \le c_0 \exp( c_1
  \paramrad)$. The result~\eqref{eqn:logistic-lower-bound} recovers this
  lower bound while applying
  whenever $\perturbsize > 0$.
\end{example}

To show some of the generality of our approach, we consider
other exponential family models. The first is similar to
the previous example.

\begin{example}[Geometric distributions]
  \label{example:geometric}
  We say $Y \sim \geometric(\lambda)$ for some $\lambda \in (0, 1)$ if $Y$
  has support $\{0, 1, 2, \ldots,\}$ and $P(Y = y) = \lambda (1 -
  \lambda)^y$.  We model this via
  $Y \mid x \sim \geometric(e^{\theta^T x} / (1 + e^{\theta^T x}))$,
  giving losses
  \begin{equation*}
    \logloss(p_\theta(\cdot \mid x), y)
    = (y + 1) \log(1 + \exp(\theta^T x)) - \theta^T x
    ~~ \mbox{and} ~~
    \scalarloss(t, y) = (y + 1) \log(1 + e^t) - t.
  \end{equation*}
  We perform a quick sketch, letting
  $b = \datarad \paramrad$ for shorthand, assuming that
  $b \ge 1$ and that $\diam(\mc{Y}) \defeq
  \max\{y \in \mc{Y}\}$ is finite
  and at least $3$.

  First, we construct a lower bound on $\qworst(t, y)$: take
  $y = \max\{y \in \mc{Y}\}$ to be the maximum element of $\mc{Y}$,
  and set $y_0 = y$ and $t = -b$. Then setting $\alpha = -1/b$ in
  the definition~\eqref{eqn:qworst}  we obtain
  \begin{align*}
    \qworst(t, y)
    \ge \frac{-\frac{y + 1}{b} \frac{1}{1 + e} + \frac{1}{b}}{
      -\frac{y + 1}{b} \frac{1}{1 + e} + \frac{1}{b}
      - (y + 1) \frac{e^b}{1 + e^b} + 1}
    = \frac{\frac{y + 1}{1 + e} - 1}{
      \frac{y + 1}{1 + e} - b + 1
      + b (y + 1) \frac{e^b}{1 + e^b}} \gtrsim \frac{1}{b}
  \end{align*}
  as $b \ge 1$ and $y \ge 3$. Additionally, we have
  $|\scalarloss'(t, y)| \gtrsim y$ and $\scalarloss''(t, y) \lesssim
  y e^{-b}$, and so, as in the derivation in
  Example~\ref{example:logistic} and by setting $\delta \gtrsim b$, we obtain that there exist numerical constants $c_0, c_1 > 0$ such that
  \begin{equation*}
    \linearconstfull \ge c_0
    y \min\left\{\sqrt{\perturbsize b}, \frac{e^{c_1 b}}{\sqrt{n}}\right\}.
  \end{equation*}
Substituting, we obtain
  the analogue of inequality~\eqref{eqn:logistic-lower-bound},
  that is,
  \begin{equation*}
    \minimax_n(\Theta, \playdists, \perturbsize)
    \ge c_0 \diam(\mc{Y}) \min\left\{
    \frac{\sqrt{\gamma \datarad \paramrad}}{\sqrt{n}},
    \frac{\exp(c_1 \datarad \paramrad)}{n} \right\}.
  \end{equation*}
  Again, we see that until $n \gtrsim \exp(c \datarad \paramrad)$, any
  method playing the models $p_\theta$ for points $\theta \in \Theta$
  necessarily cannot converge faster than $\diam(\mc{Y}) / \sqrt{n}$.
\end{example}

Poisson regression yields a similar lower bound:

\begin{example}[Poisson regression]
  \label{example:poisson}
  In the poisson regression problem, we model $y \in \N$ as
  $\poisson(e^{\theta^T x})$, so that
  \begin{equation*}
    -\log p_\theta(y \mid x) = e^{\theta^T x} - y\theta^T x + \log(y !),
  \end{equation*}
  and we may consider the loss $\scalarloss(t, y) = e^t - yt$.
  We claim that in the setting of Theorem~\ref{thm:lowerbound},
  where we set $\diam(\mc{Y}) = \max\{y \in \mc{Y}\} \ge 3$ as in
  Example~\ref{example:geometric}, we have
  \begin{equation}
    \label{eqn:poisson-lower}
    \minimax_n(\Theta, \playdists, \perturbsize)
    \ge c_0 \min\left\{\diam(\mc{Y})
    \frac{\sqrt{\perturbsize \datarad \paramrad}}{\sqrt{n}},
    \frac{e^{c_1 \datarad \paramrad} \diam(\mc{Y})^2}{n} \right\}.
  \end{equation}

  To see the lower bound~\eqref{eqn:poisson-lower}, it is sufficient to
  lower bound the linearity constant, and we may assume that $\datarad
  \paramrad \ge 2$. Our first step is to lower bound the quantity
  $\qworst$. Let $b = \datarad \paramrad / 2$ for shorthand, and set $t =
  b$, $y = \diam(\mc{Y})$, and $y_0 = y$, and $\alpha = -\frac{1}{b} \ge -1$,
  so that $\scalarloss'(t, y) = e^t - y = e^{-b} - y < 0$ while
  $\alpha \scalarloss'(\alpha t, y) = -\frac{1}{b}(e - y) > 0$.
  As a consequence, completely parallel to our derivation
  in Example~\ref{example:geometric}, we obtain $\qworst(t, y)
  \gtrsim \frac{1}{b}$. Moreover, the bounds
  $\scalarloss''(t, y) = e^t$ and the choice $\delta = b/2$ in
  the definition~\eqref{eqn:def-linearconst} yield
  that for numerical constants $c_0, c_1 > 0$ we have
  \begin{equation*}
    \linearconstfull \ge c_0 y \min\left\{\sqrt{\perturbsize b},
    \frac{e^{c_1 b} y}{\sqrt{n}} \right\}.
  \end{equation*}
  Substituting this into Theorem~\ref{thm:lowerbound} then implies the
  claim~\eqref{eqn:poisson-lower}.
\end{example}

In contrast, for linear regression problems, there is limited
worst-case behavior, which is natural: the problem is always strongly
convex, no matter the misspecification of the model.

\begin{example}[Linear regression]
  \label{example:linear-regression}
  Again we consider the log loss, but we assume
  that our model is that
  $y \mid x \sim \normal(\theta^T x, 1)$, so that the loss
  becomes
  $\logloss(p_\theta(\cdot \mid x), y)
  = \frac{1}{2} (\theta^T x - y)^2$ and
  $\scalarloss(t, y) = \frac{1}{2}(t - y)^2$.
  In this case, we may take $\qworst \gtrsim 1$ in Eq.~\eqref{eqn:qworst},
  and the linearity constant~\eqref{eqn:def-linearconst} becomes
  \begin{align*}
    \linearconstfull
    & \asymp
    \sup_{y \in \mc{Y}}
    \sup_{t^2 + \delta^2 \le \frac{\datarad^2 \paramrad^2}{2}}
    \left\{(t - y)
    \min\left\{\delta \sqrt{\perturbsize},
    \frac{t - y}{\sqrt{n}} \right\} \right\} \\
    & \asymp
    \min\left\{ \datarad \paramrad
    \max\{\datarad \paramrad,
    \diam(\mc{Y})\}
    \sqrt{\perturbsize},
    \frac{\max\{\datarad^2 \paramrad^2, \diam(\mc{Y})^2\}}{\sqrt{n}}
    \right\},
  \end{align*}
  yielding minimax lower bound
  \begin{equation*}
    \minimax_n(\Theta, \playdists, \perturbsize)
    \ge c \min\left\{\frac{\max\{\datarad^2 \paramrad^2,
      \datarad \paramrad \diam(\mc{Y})\} \sqrt{\perturbsize}}{\sqrt{n}},
    \frac{\max\{\datarad^2 \paramrad^2, \diam(\mc{Y})^2\}}{n}
    \right\}.
  \end{equation*}
  This is sharp: the stochastic gradient method achieves the
  bound~\cite{HazanKa11}, as $\scalarloss$ is strongly convex and has
  Lipschitz constant $\max\{\paramrad \datarad, \diam(\mc{Y})\}$.

  With that said, this behavior---which depends on $\diam(\mc{Y})$---is
  worse than what one achieves in well-specified or stochastic
  settings, where the stochasticity means that rates of $1/n$ are achievable.
\end{example}

\subsection{Scoring rules and general losses}
\label{sec:general-losses}

In probabilistic prediction and forecasting, one more generally may consider
\emph{scoring rules}~\cite{GneitingRa07}, which are losses designed to
engender various behaviors: honesty in eliciting predictions, calibration of
forecasts, robustness, or other reasons. Typically, these induced losses are
exp-concave (as we discuss in the next section, which will allow us to
describe an efficient algorithm for them). For example, to achieve
robustness~\cite{MeiBaMo18, HuberRo09} (there are no unbounded losses) one
might consider the squared error
or Hellinger-type losses
\begin{equation}
  \label{eqn:square-hellinger-loss}
  \losssq(p(\cdot \mid x), y) \defeq \half
  (p(y \mid x) - 1)^2
  ~~ \mbox{and} ~~
  \losshel(p(\cdot \mid x), y) =
  (\sqrt{p(y \mid x)} - 1)^2,
\end{equation}
neither of which is proper (so that the true distribution may not minimize
the loss). Alternatively, proper scoring rules~\cite{GneitingRa07} are
minimized by the true predictive distribution, and include the logarithmic
loss and the quadratic scoring rule with loss
\begin{equation}
  \label{eqn:quadratic-loss}
  \lossquad(p(\cdot \mid x), y)
  \defeq \half \sum_{k \in \mc{Y}} (p(k \mid x) - \indic{k = y})^2
  = \half (p(y \mid x) - 1)^2 + \half \sum_{k \neq y} p(k \mid x)^2.
\end{equation}
As these scoring rules are differentiable and at least $\mc{C}^2$ on $p \in
(0, 1)$, an argument by the delta method~\cite{VanDerVaart98} shows that in
well-specified cases, one expects to achieve convergence at rate $1/n$.
Moreover, as we will see in the next section, aggregating algorithms can
achieve regret scaling as $\log n$ for each of these loss measures.

\newcommand{\rate}{\mathsf{r}(n)}

Yet, at least in an asymptotic sense, the minimax bound in
Theorem~\ref{thm:lowerbound} shows that this is unachievable with
misspecification. Here, because of the complexity of the losses and
resulting calculations, we take a completely asymptotic perspective, saying
that we have an \emph{asymptotic rate $\rate$ minimax lower bound} if $\rate
\to 0$ as $n \to \infty$, while
\begin{equation*}
  \liminf_{n \to \infty} \frac{\minimax_n(\Theta, \playdists, \perturbsize)}{
    \rate} > 0
\end{equation*}
for all $\perturbsize > 0$.
Each of the bounds in
Examples~\ref{example:logistic}--\ref{example:linear-regression} is then
asymptotic rate $\rate = \frac{1}{n}$.  Once we move beyond the log loss to
alternative scoring rules, however, the rate $\rate = \frac{1}{n}$ is
no longer achievable with misspecification if $\playdists = \{p_\theta\}$
are the proper models.

As usual we consider losses taking the form
$\loss(p_\theta(\cdot \mid x), y)
= \scalarloss(\theta^T x, y)$ for some scalar
induced loss $\scalarloss : \R \times \mc{Y} \to \R$.
We restrict our focus to losses for which no \emph{universally
perfect} prediction exists, meaning that
if $t \in \R$ and $y \in \mc{Y}$ satisfy
$\scalarloss'(t, y) \neq 0$, there exists $y_0 \in \mc{Y}$
such that $\scalarloss'(t, y) \scalarloss'(t, y_0) < 0$.
We have the following result,
whose proof
we provide in Appendix~\ref{sec:proof-slow-rates-oops}.
\begin{proposition}
  \label{proposition:slow-rates-oops}
  Assume that the scalar loss allows no universally perfect prediction.  If
  there exists $|t| \le \datarad \paramrad/2$ and $y \in \mc{Y}$ such that
  $\scalarloss'(t, y) \neq 0$ while $\scalarloss''(t, y) = 0$ and
  $\scalarloss(\cdot, y)$ is $\mc{C}^3$ near $t$, then the prediction family
  $\playdists = \{p_\theta\}_{\theta \in \Theta}$ has asymptotic rate $\rate
  = n^{-3/4}$ minimax lower bound.
\end{proposition}
\noindent
Roughly, the result in the proposition is simple: if the
induced scalar loss $\scalarloss$ is not convex in $t$, then
proper predictions cannot be rate-optimal (as rates scaling as
$1/n$ are achievable here).

While it is possible in some cases
to achieve explicit constants---for example, for logistic
regression this is relatively straightforward---in general it is somewhat
tedious. Nonetheless, we have the following result,
whose tediousness in verification precludes our including a formal
proof, but essentially we need simply note that the
induced losses $\scalarloss(t, y)$ for each problem are non-convex in $t$
but are smooth.
\begin{corollary}
  Let $\playdists = \{P_\theta\}$ be any of the logistic-,
  geometric-, poisson-, or linear-regression families of
  predictive densities. Then for any of the squared $\losssq$, Hellinger
  $\losshel$, or quadratic $\lossquad$ losses
  (Eqs.~\eqref{eqn:square-hellinger-loss} and \eqref{eqn:quadratic-loss}),
  $\minimax_n(\Theta, \playdists, \perturbsize)$ has asymptotic rate
  lower bound  $\rate
  = \frac{1}{n^{3/4}}$ for any $\perturbsize > 0$.
\end{corollary}

Theorem~\ref{thm:lowerbound} and its consequences via
Proposition~\ref{proposition:slow-rates-oops} assert that \emph{any}
algorithm returning elements of the parametric family $\pt$ must suffer when
misspecification is possible. These results are information-theoretic, and
as such, we see that in situations where misspecification is possible, to
perform better it is essential that the family $\playdists$ of allowable
distributions be improper.



%% file: sections/upperbound.tex

\section{Robustness via Improper Learning}
\label{section:upperbound}

While proper algorithms evidently must suffer losses when they are
misspecified, we now show that simply by considering $\Gamma = \conv \pt$
we can sidestep the lower bound of
Theorem~\ref{thm:lowerbound}. Specifically, we consider Vovk's Aggregating
Algorithm and show that this provides stability to misspecification. We
present the algorithm in an online setting, though standard online-to-batch
conversion techniques~\cite{CesaBianchiLu06} extend the result to the
stochastic optimization setting in which we have proved each of our
lower bounds.

\begin{algorithm} [h]
  \caption{\label{alg:vovk} Vovk's Aggregating Algorithm (Online Setting)}
  Define $\Gamma = \conv \param$ and let $\eta$ be some fixed value $>0$. \\
  For $t = 0, \ldots, n$ \\
  \hspace*{10mm} Nature reveals $x_{t}$. \\
  \hspace*{10mm} Define $ d \vovkmu[t] (\theta) \propto
  \exp(- \eta \sum_{s = 0}^{t-1} \Loss(P_{\theta}(y_{s} \mid x_{s}))).$ \\
  \hspace*{10mm} Decision Maker plays
  $\vovkP[t] \defeq \int_{\theta \in \Theta} P_{\theta}(\cdot \mid x_{t})
  d \vovkmu[t] (\theta)$. \\
  \hspace*{10mm} Nature reveals $y_{t}$ and Decision Maker suffers loss
  $\Loss(\vovkP[t](y_t \mid x_t))$.
\end{algorithm}

\begin{table}[ht]
  \begin{center}
    \begin{tabular} {c c c c c}
      \toprule
      & $\logloss$
      & $\losssq$
      & $\losshel$
      & $\lossquad$ \\
      \hline
      Mixability Constant $\eta$  & $1$ & $1$ & $3$ & $1/2$\\
      \bottomrule
    \end{tabular}%
  \end{center}
  \caption{\label{table:examples} Mixability constants for common losses
    in Eqs.~\eqref{eqn:quadratic-loss} and
    Eqs.~\eqref{eqn:square-hellinger-loss}, where
    $\logloss(p, y) = -\log p(y)$,
    $\losssq(p, y) = \half (p(y)-1)^2$,
    $\losshel(p, y) = (\sqrt{p(y)}-1)^2$,
    and $\lossquad(p, y) = \half\ltwo{p(\cdot) - e_y}^2$.    
    See Appendix~\ref{append:mix}.}
\end{table}

To give a regret bound, we continue our usual focus by considering losses
$\loss$ and families $\pt$ for which we can write $\loss(p_\theta(\cdot \mid
x), y) = \scalarloss(\theta^T x, y)$. We restrict ourselves somewhat to
considering \emph{mixable} losses, where for some $\eta > 0$, the function
$p \mapsto \exp(-\eta \loss(p, y))$ is concave over the collection
$\mc{P}(\mc{Y})$ of distributions on $\mc{Y}$.  This constant $\eta$
guaranteeing the exp-concavity of $\loss$ bounds the \emph{mixability}
constant, which allows one to obtain ``fast rates'' via
exponentially-weighted averaging in many online learning
problems~\cite{Vovk98, CesaBianchiLu06, VanErvenGrMeReWi15}. In
Table~\ref{table:examples}, we record the mixability constants for several
example losses---each of which we touch on in
Sec.~\ref{sec:general-losses}---showing that Vovk's aggregating algorithm
achieves logarithmic regret for any of them once we apply the coming
convergence result.
\begin{corollary}[Foster et al.~\cite{FosterKaLuMoSr18}, Theorem 1]
  \label{corollary:bayes}
  Let $\what{P}_{t, \eta}^{\mathsf{Vovk}}$ be as defined above. Let
  $\lip(T)$ be the Lipschitz constant of $\ell$ restricted to $[-T, T]
  \times \mc{Y}$. Then for any sequence $(x_i, y_i)_{i=1}^n$ and
  $\theta\opt \in \Theta$,
  \begin{align*}
    \regret(\what{P}_{n, \eta}^{\mathsf{Vovk}}, p_{\theta\opt}) \leq
    5 \frac{d}{\eta} \log\left(\frac{\lip(\datarad \paramrad)n }{d} + e \right).
  \end{align*}
\end{corollary}

Using Corollary~\ref{corollary:bayes} and Theorem~\ref{thm:lowerbound}, we
see that the aggregating algorithm typically provides a stronger convergence
guarantee than algorithms constrained to $\pt$ can attain. In each of
Examples~\ref{example:logistic}--\ref{example:poisson}, the lower bounds
necessarily suffer exponential dependence $\exp(\Omega(1) \datarad
\paramrad)$ as $n \to \infty$, and so as long as the Lipschitz constant
$\lip(\datarad \paramrad)$ is not super-exponential in $\datarad \paramrad$,
Corollary~\ref{corollary:bayes} guarantees better convergence. Even more, in
some cases---for example, when using the general (potentially non-convex in
$\theta$) losses as in Sec.~\ref{sec:general-losses}---even for fixed radii
$\datarad, \paramrad$ we have $\sqrt{n} \linearconstfull \to \infty$ as $n
\to \infty$. In this case, Corollary~\ref{corollary:bayes} even guarantees a
better asymptotic rate in $n$.

\subsection{Lower bounds for arbitrary improper algorithms}

\newcommand{\allplaydists}{\playdists_{\textup{all}}}

An important question is whether the Aggregating Algorithm~\ref{alg:vovk}
achieves optimal rates when the misspecification parameter $\perturbsize$
changes. As the coming Theorem~\ref{theorem:well-specified-lower}
shows, Corollary~\ref{corollary:bayes} is
tight to within logarithmic factors, as we can show an (asymptotic) lower
bound of $d/n$ even for well-specified families of generalized linear
models. The theorem also shows that while playing in the
convex hull of a parameterized family $\param$ allows more powerful
mechanisms, in natural (well-specified) scenarios, this extra power is no
panacea: the upper bound on the risk of the Aggregating
Algorithm~\ref{alg:vovk} is tight to a logarithmic factor over all
algorithms that may play arbitrary elements in the convex hull (or even
algorithms playing any probability distribution).
Indeed, let $\allplaydists$ be the collection of
\emph{all} probability distributions on $Y \mid X$, so that an algorithm
may play an arbitrary distribution (which is of course
larger than $\conv \pt$). We then take the risk to be the expected logarithmic
loss, where for a conditional distribution $p(y \mid x)$ we define
\begin{equation*}
  \risk_P(p) \defeq \E_P[\logloss(p(\cdot \mid X), Y)]
  = -\E_P[\log p(Y \mid X)],
\end{equation*}
and we let $\risk_P\opt = \inf_p \risk_P(p)$ be the smallest
risk across all predictive distributions $p(y \mid x)$.
When the models $p_\theta$ are not misspecified, i.e.,
$\perturbsize = 0$, we have $\risk_{P_\theta}\opt = \risk_{P_\theta}(p_\theta)$,
and we have the following lower bound.

\begin{theorem}
  \label{theorem:well-specified-lower}
  Let $\param$ be a generalized linear model of the form 
  \begin{equation*}
    dP_{\theta}(y \mid x ) = \exp(y \theta^{T}x - A(\theta^{T}x))
    d\nu(y),
  \end{equation*}
  where $\nu$ is a base measure and $X \sim \uniform(\{-1, 1\}^d)$,
  $0$ is in the interior of $\dom A$, and let
  $\Theta$ contain $\zeros$ in its interior.  Then for the log loss
  $\logloss$, there exists a numerical constant $c > 0$ such that for all
  large enough $n$
  \begin{equation*}
    \minimax_n(\Theta, \allplaydists, 0)
    = \inf_{\what{p}_n} \sup_{\theta \in \Theta}
    \E_{P_\theta}^n\left[\risk_{P_\theta}(\what{p}_n)
      - \risk_{P_\theta}\opt \right]
    \ge c \frac{d}{n}.
  \end{equation*}
\end{theorem}
\noindent
See Appendix~\ref{appendix:optseq} for the proof of Theorem
\ref{theorem:well-specified-lower}. As we essentially only
care about the conditional distribution $p_\theta(y \mid x)$, here
we chose the marginal over $X$ to be uniform for convenience; other choices
suffice as well.

Comparing the lower bound Theorem~\ref{theorem:well-specified-lower}
provides with the regret bound in Corollary~\ref{corollary:bayes}, we see
that holding $\datarad \paramrad$ constant (and $\lip(\datarad
\paramrad)$ constant) then for risk functional $\risk_P(p)
\defeq \E_P[\logloss(p(\cdot \mid X), Y)]$,
a standard online-to-batch conversion (or Jensen's inequality) implies
\begin{equation*}
  \E\left[\risk_P(\what{p}_{n}^{\mathsf{Vovk}}) - \inf_{\theta\opt \in \Theta}
    \risk_P(p_\theta\opt)\right]
  \le (5 + o(1)) \frac{d \log n}{n}
\end{equation*}
as $n \to \infty$, where $o(1) \to 0$ hides problem-dependent constants.
To within a factor $O(1) \log n$, then, Vovk's aggregating algorithm
is generally unimprovable.

\subsection{Asymptotically aggregating to point predictions}
\label{section:asymp}

While in general the ability to play elements in $\conv \param$ could (in
principle) yield much better performance than any individual element
$p_\theta$ for $\theta \in \Theta$, in a sense, the aggregating algorithm is
only performing a small amount of averaging to substantially increase its
robustness. Indeed, when the risk minimization problem at hand is
classical---the loss has continuous derivatives and the population risk
$\risk_P$ is strongly convex in a neighborhood of its minimizer
$\theta\opt$---then we can show that Vovk's aggregating algorithm
asymptotically plays points very close to $\pt$. That is, in ``nice'' cases,
the aggregating algorithm more or less behaves as the empirical risk
minimizer, which is asymptotically optimal~\cite{DuchiRu20}.  In these
cases, stochastic gradient methods (which are necessarily proper, as they
optimizer over $\theta$) similarly achieve optimal asymptotic
rates~\cite{DuchiRu20}, and sometimes similarly strong finite sample
rates~\cite{Bach14}.

We make this more formal via a generalized Bernstein von-Mises
Theorem, which shows that when a unique minimizer exists, the density $d
\vovkmu$ in Alg.~\ref{alg:vovk} converges to a normal density centered at
the empirical minimizer $\empmin \in \Theta$ with covariance operator
shrinking at the rate $1/n$.  Such posterior limiting
normality results are relatively well-known: see
\citep{VanDerVaart98, KleijnVa12}. In our setting, rather than considering
just the posterior distribution---as our models may be misspecified, so
that a posterior is less sensible---we consider distributions over
$\Theta$ of the form of $\vovkmu$; when $\Loss$ is the log-loss, this is the
usual posterior. We first define the class of families and losses for which
Theorem \ref{thm:bvm} holds.

\begin{definition}
  \label{def:bvmclass}
  The family and loss pair $(\param, \loss)$ is $\bvmgeneral$ if there exist
  $\epsilon_1, \epsilon_2 > 0$ such that the risk
  $\risk_P(\theta) \defeq \E_{P}[\Loss(P_{\theta}(Y \mid X))]$ satisfies the
  following conditions:
  \begin{enumerate}[(i)]
  \item The minimizer $\theta\opt = \argmin_{\theta
    \in\Theta} \risk_P(\theta)$ is unique and has positive
    definite Hessian $\nabla^2
    \risk_P(\theta\opt) \succ 0$.
  \item \label{item:local-lipschitz}
    On the $\epsilon_1$-ball around $\theta\opt$, $\theta\opt +
    \epsilon_1 \ball_2^d$, the loss $\theta \mapsto \loss(P_\theta(\cdot
    \mid x), y)$ is $\lipobj(x, y)$ Lipschitz and has $\liphess(x,
    y)$-Lipschitz Hessian, where $\E[\lipobj(X, Y)^2] < \infty$ and
    $\E[\liphess(X, Y)] < \infty$.
  \item \label{cond:identifiability}
    For all $\theta \in \Theta \setminus \{\theta\opt + \epsilon_1
    \ball_2^d\}$, we have $\risk_P(\theta) \ge \risk_P(\theta\opt) + \epsilon_2$.
  \end{enumerate}
\end{definition}
\noindent
When $\theta \mapsto \loss(P_\theta(\cdot \mid x), y)$ is convex,
condition~\eqref{cond:identifiability} is redundant given the others,
and the other conditions of Definition~\ref{def:bvmclass} hold for
generalized linear models.
Under the conditions Definition~\ref{def:bvmclass} specifies,
we then obtain the following convergence guarantee, whose
proof we provide in Appendix~\ref{appendix:bvm}.

\begin{theorem} [Generalized Bernstein von-Mises]
  \label{thm:bvm}
  Let the pair $(\param, \loss)$ be Bernstein von-Mises generalizable
  (Definition~\ref{def:bvmclass}), and for $(X_i, Y_i) \simiid P$ define
  $\risk_n(\theta) = \frac{1}{n} \sum_{i=1}^{n} \Loss (P_{\theta}(\cdot \mid
  X_i), Y_i)$. Assume $\Theta$ is compact and $\theta\opt \in \interior
  \Theta$. Let $\empmin \defeq \argmin_{\theta \in \Theta} \risk_n(\theta)$
  and $\vovkmu$ be as defined in Vovk's Aggregating Algorithm. Then
  \begin{equation*}
    \tvnorm{\vovkmu
      - \normal\left(\empmin, \frac{1}{n} \nabla^2 \risk_n(\empmin)^{-1}
      \right)}
    \mathop{\longcas}_{P} 0.
  \end{equation*}
\end{theorem}

Using the theorem and its proof, we can also (under a minor
continuity condition) establish a convergence guarantee showing
roughly that the aggregating algorithm asymptotically
plays essentially the empirical point estimator.
We consider the following assumption.

\begin{assumption}
  \label{assumption:extra}
  There exists a neighborhood $B$ of
  $\theta\opt$ such that
  the log-likelihood $\theta \mapsto \log p_\theta(y \mid x)$ is
  $\lipP(x, y)$-Lipschitz  on $B$, and
  $\lipP(x) \defeq \sup_{\theta \in B} \int_{\mc{Y}} \lipP(x, y) dP_\theta(y \mid x)
  < \infty$ for each $x$.
\end{assumption}

\noindent
We then have the following corollary,
whose proof we provide in Appendix~\ref{appendix:corbvm}.
\begin{corollary}
  \label{cor:bvm}
  In addition to the conditions of Theorem~\ref{thm:bvm}, let
  Assumption~\ref{assumption:extra} hold.
  Then for each $x \in \mc{X}$,
  \begin{equation*}
    \tvnorm{\vovkP(\cdot \mid x) - P_{\empmin}(\cdot \mid x)}
    \cas 0.
  \end{equation*}
\end{corollary}

Roughly, Theorem~\ref{thm:bvm} and Corollary~\ref{cor:bvm} show that
the aggregating algorithm~\ref{alg:vovk} is asymptotically
constrained to making predictions in $\pt$,
at least in non-adversarial cases. In a sense, then,
the aggregating algorithm~\ref{alg:vovk} is not taking
full advantage of its improperness: while it can
return any distribution in $\conv \pt$, it (eventually) is
nearly playing elements of $\pt$. While this is
optimal in some cases (Theorem~\ref{theorem:well-specified-lower}),
the question of how to efficiently and optimally return predictions
in $\conv \pt$ remains open and a natural direction for future work.

%% file: sections/experiments.tex

\newcommand{\Pmix}{P^{\mathsf{mix}}_{n,\paramrad}}

\section{Experiments and Implementation Details}
\label{section:experiments}

Before discussing our experiments, we make a few remarks on 
the computability of Vovk's Aggregating Algorithm. Whenever
$\Loss(P_{\theta}(\cdot \mid x), y)$ is convex in $\theta$ and
$\beta$-smooth, there exists an algorithm~\cite{FosterKaLuMoSr18}
approximating $\vovkP$ that achieves the regret bound in
Corollary~\ref{corollary:bayes} to within an additive factor $1/n$, and the
algorithm is polynomial in $(\datarad \paramrad, d, \lip(\datarad
\paramrad), n)$. Yet these algorithms are still computationally intensive;
assuming our theoretical results are predictive of actual performance,
one might expect that aggregating-type strategies could still yield
improvements over standard empirical risk minimization.
Indeed, \citet{JezequelGaRu20} take the computational difficulty of
the approximating algorithms in the paper~\cite{FosterKaLuMoSr18}
as motivation to develop an efficient
improper learning algorithm for the special case of logistic
regression, which (roughly) hedges its predictions by pretending to receive
both positive and negative examples in future time steps, constructing
a loss that depends explicitly on the new data $x_t$;
\citeauthor{JezequelGaRu20} show that it achieves a regret
bound with a multiplicative $\datarad \paramrad$ factor of
the logarithmic regret in Corollary~\ref{corollary:bayes}.
It is unclear how to extend this approach to situations in which the
cardinality $|\mc{Y}|$ of $\mc{Y}$ is much larger than 1, though this
is an interesting question for future work.
In our experiments, we take a heuristic approach, focusing on the risk
minimization setting, and perform aggregation of subsampled maximum
likelihood estimators; this approach is reminiscent of the subsampled and
bootstrapped estimators~\cite{ZhangDuWa13, KleinerTaSaJo12}, but
we use aggregation as in Alg.~\ref{alg:vovk} to weight predictions.
We call the procedure
AHA (\textbf{A} \textbf{H}euristic \textbf{A}ggregation) for short.

\begin{algorithm} [h]
  \caption{\label{alg:aha}
    AHA (\textbf{A} \textbf{H}euristic \textbf{A}ggregation)}
  Input: $\{ (x_i,y_i) \}_{i=1}^n $ and parameter radius $\paramrad$ \\
  Output: $\Pmix$ \\
  \hspace*{5mm} \textbf{For $k = 1, \cdots, K$ } \\
  \hspace*{10mm} $S_k \gets$ random subset of the data of size $|S_k| = 2n/3$ \\
  \hspace*{10mm} $\what{\theta}^{k}_n \gets \argmin_{\norm{\theta} \leq B} \sum_{(x,y) \in S_k} L(p_{\theta}(y \mid x) )$ \\
  \hspace*{10mm} $\mu^{k}_n \gets \exp(\sum_{i=1}^n
  -L(p_{\what{\theta}^k_n}(y_i \mid
  x_i) ))$ \\
  \hspace*{5mm} $\Pmix \gets \big( \sum_{k=1}^K \mu_n^k p_{\what{\theta}^{k}_n}\big) / \big( \sum_{k=1}^K \mu_n^k \big)$
\end{algorithm}

Our results suggest that when performing a probabilistic forecasting task
with parameterized model $\param$, returning the mixture distribution from
Vovk's Aggregating Algorithm should be more robust to misspecification than
an algorithm which returns $P_{\what{\theta}}$; we thus expect
Algorithm~\ref{alg:aha} should exhibit more robustness as well. To that end,
we consider two experiments: the first a synthetic experiment with linear
regression, where we may explicitly control the degree of misspecification,
and the second a logistic regression problem on real digit recognition data,
where we mix two populations and we expect (roughly) that a model should do
well on one, but may be missing important aspects of the other.

\paragraph{Improper Linear Regression} For the
synthetic data,
we let $X \in \R^d$ be an observed covariate and $H \in \R$ a hidden
variable, and for $\tau \in \R_+$ we let $y$ have density
\begin{equation*}
  p_{\tau}(y \mid X=x, H=h) = \frac{1}{\sqrt{2 \pi}} e^{ -\frac{1}{2}(y -
    (x^T \theta\opt + \tau h)^2}.
\end{equation*}
We fix the dimension $d = 10$ and let
$\theta\opt \in \R^d$ be uniform on $\sphere^{d-1}$;
we generate data by drawing $(X, H) \sim \normal(0, I_d) \times \normal(0, 1)$.
We then use the parametric model $\pt$ to model
$Y \mid X = x \sim \normal(\theta^T x, 1)$, which is misspecified
when $\tau > 0$. As $\tau$
grows larger---increasing misspecification---we
expect greater differences between the M.L.E.\ $\empmin$ and the
AHA Algorithm~\ref{alg:aha}.

\paragraph{Logistic Regression} We consider the MNIST handwritten
digits~\cite{LeCunEtAl95},
where we mix in typed digits; as our base featurization, we use a standard
7-layer convolutional
neural network trained on the MNIST data, so that the typewritten digits
(roughly)  represent a misspecified sub-population, and as the
proportion $\tau$ of typewritten digits increases, we expect increasing
misspecification. We consider a simplified binary version of this problem,
where we seek to distinguish digits $3$ and $8$, and we use a logistic
regression model $p_\theta(y \mid x) = (1 + e^{-y x^T \theta})^{-1}$
with log loss.

\paragraph{Experiment}
For both linear regression and logistic regression, we conduct the following
experiment: For a training sample size $n$
and parameter radius $\paramrad$,
we compute the constrained MLE
\begin{equation*}
  \empmin
  \defeq \argmin_{\norm{\theta} \leq \paramrad}
  \sum_{i=1}^n \loss(P_{\theta}(\cdot \mid x_i), y_i)
\end{equation*}
and return $P_{\empmin}$, and also compute $\Pmix$ as the output of
Alg.~\ref{alg:aha} with resampling size $K = 20$ for the improper linear regression experiment and $K = 10$ for the MNIST experiment. We use a held-out
test set of size $N = 5000$ to approximate the risk $\risk(p)$ of the
returned conditional probability $p$. We plot how this approximated risk decays as we increase training sample size $n$ up to $1000$ for improper linear regression and up to $200$ for logistic regression.

\begin{figure} [h]
  \centering
  \begin{subfigure}{0.32\textwidth}
    \includegraphics[height=1.7in]{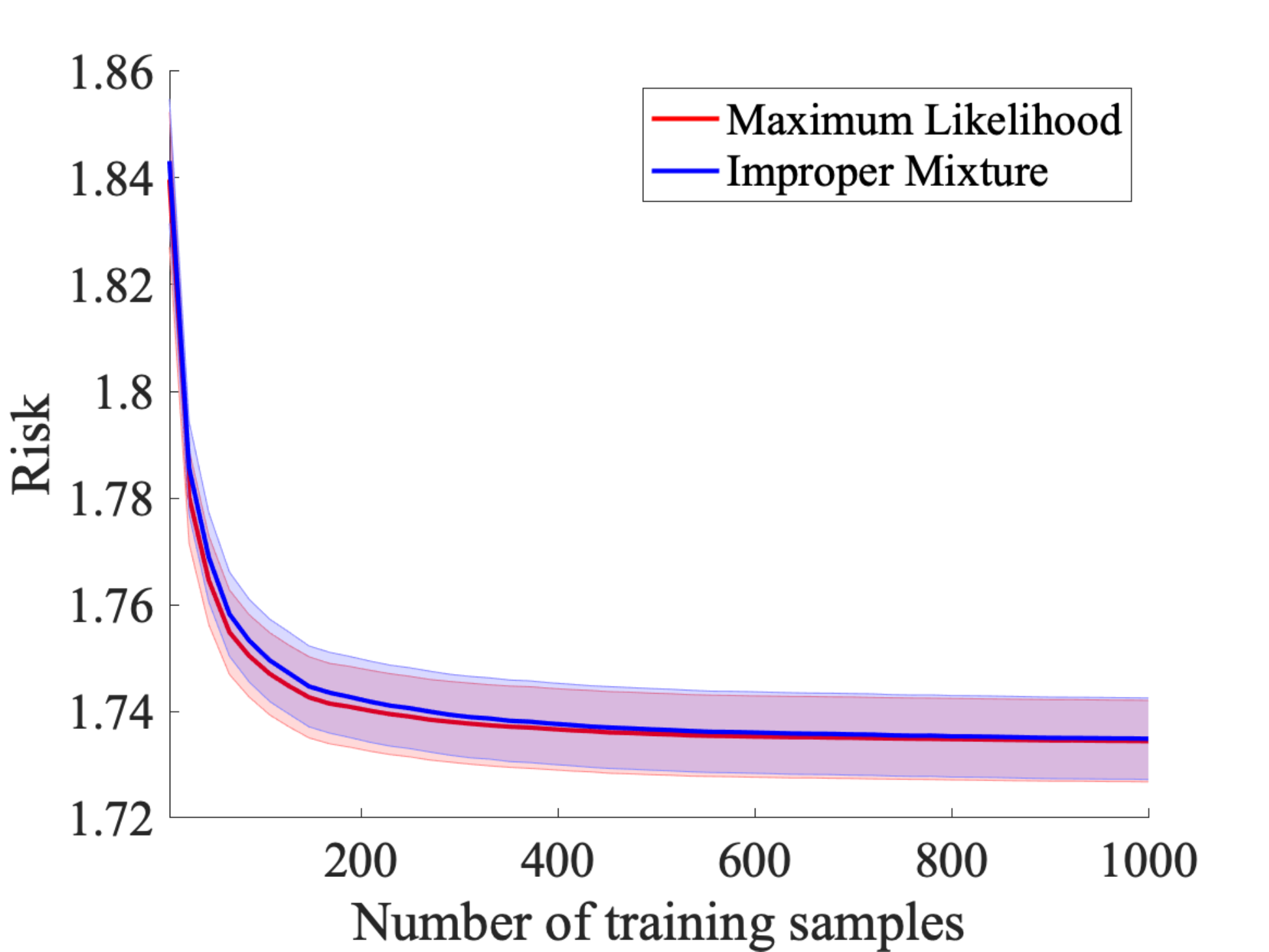} 
    \caption{$\tau = 0$}
  \end{subfigure} %
  \begin{subfigure}{0.32\textwidth}
    \includegraphics[height=1.7in]{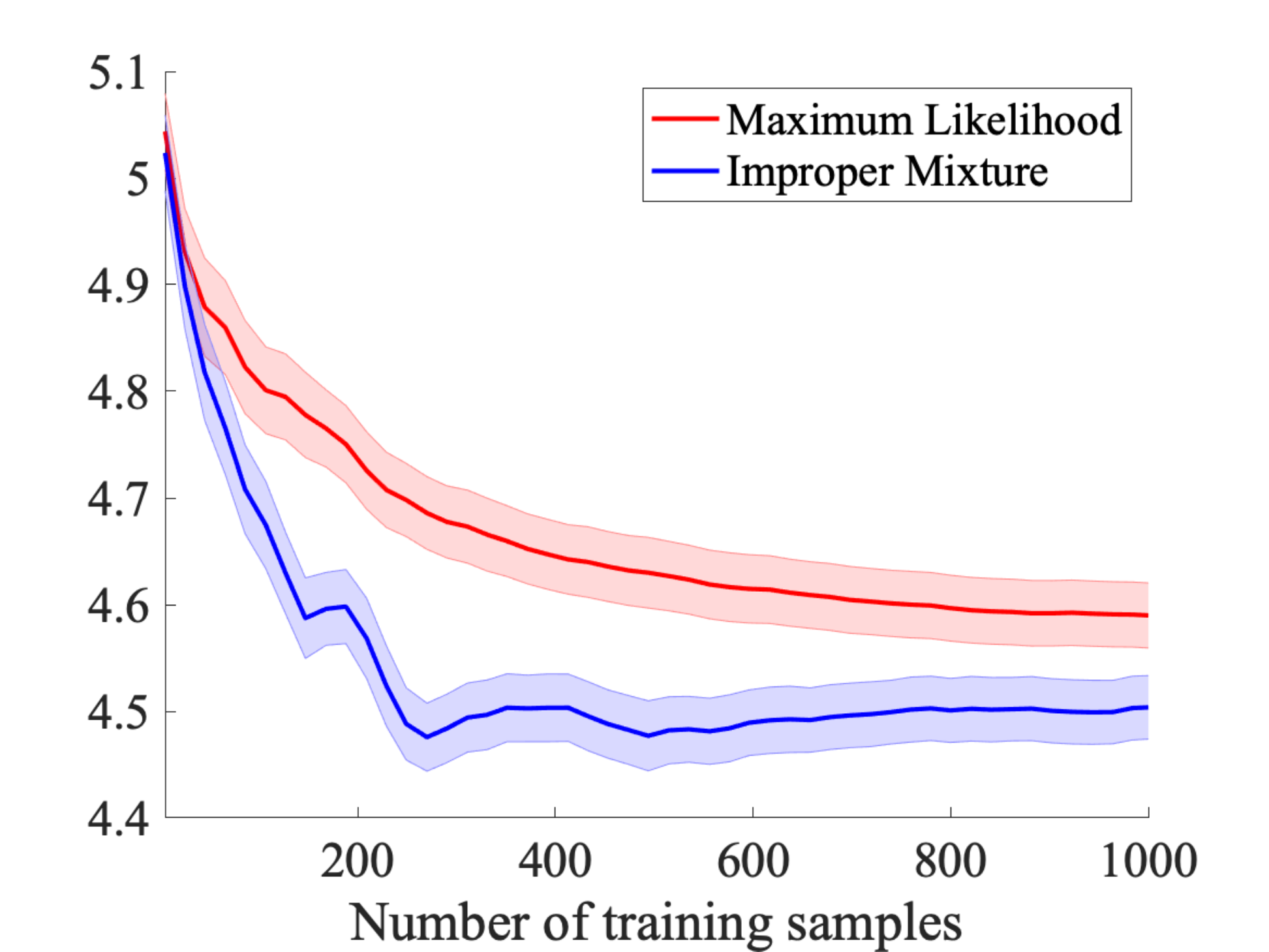} 
    \caption{$\tau = 2.5$}
  \end{subfigure} %
  \begin{subfigure}{0.32\textwidth}
    \includegraphics[height=1.7in]{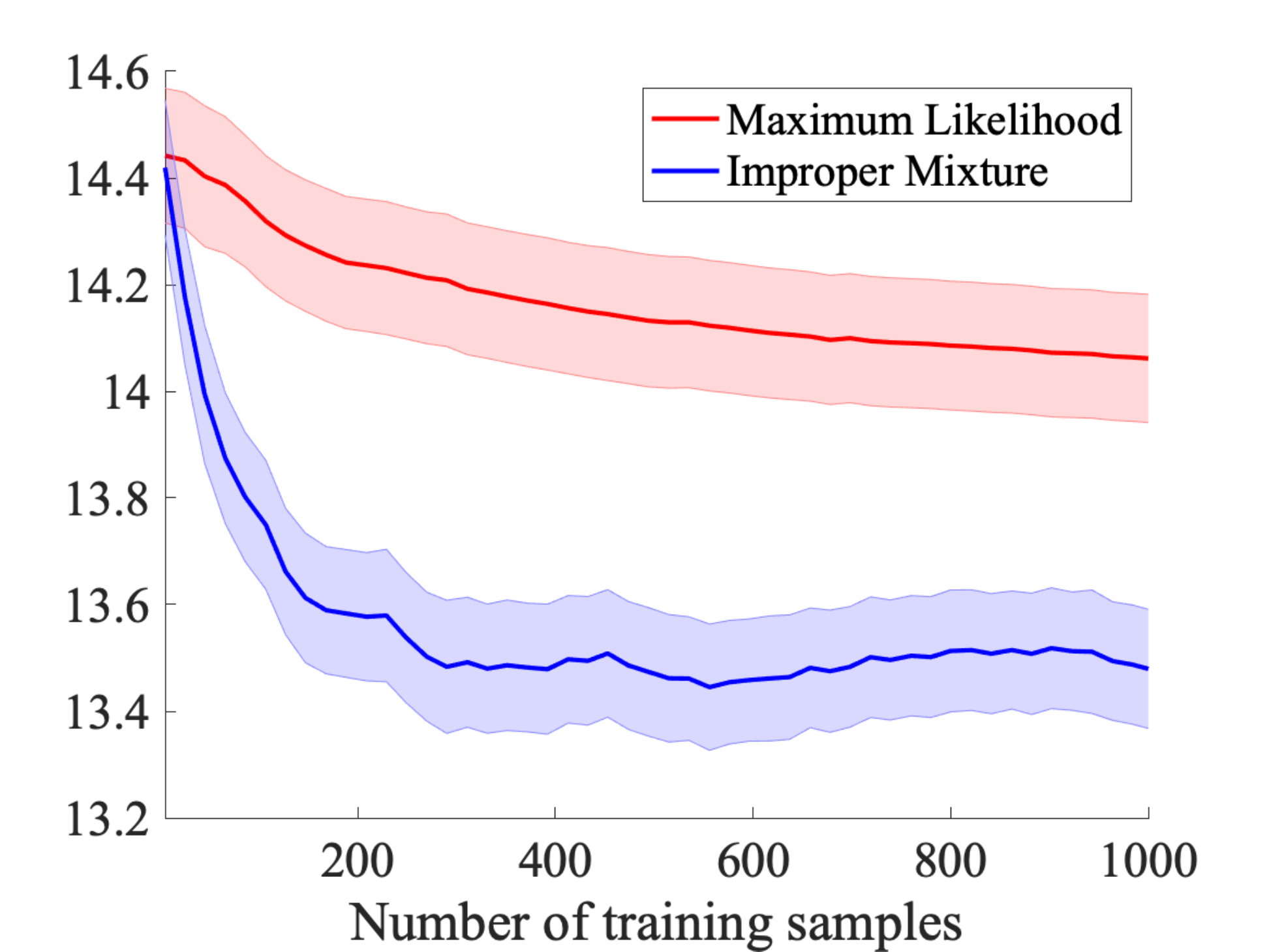} 
    \caption{$\tau = 5$}
  \end{subfigure} 
  \caption{Linear Regression, Synthetic Data. As misspecification $\tau$
    increases, the improper learning algorithm AHA (Alg.~\ref{alg:aha})
    outperforms the best constrained MLE.}
  \label{fig1}
\end{figure}

\begin{figure}[h]
\centering
\begin{subfigure}{0.32\textwidth}
  \includegraphics[height=1.7in]{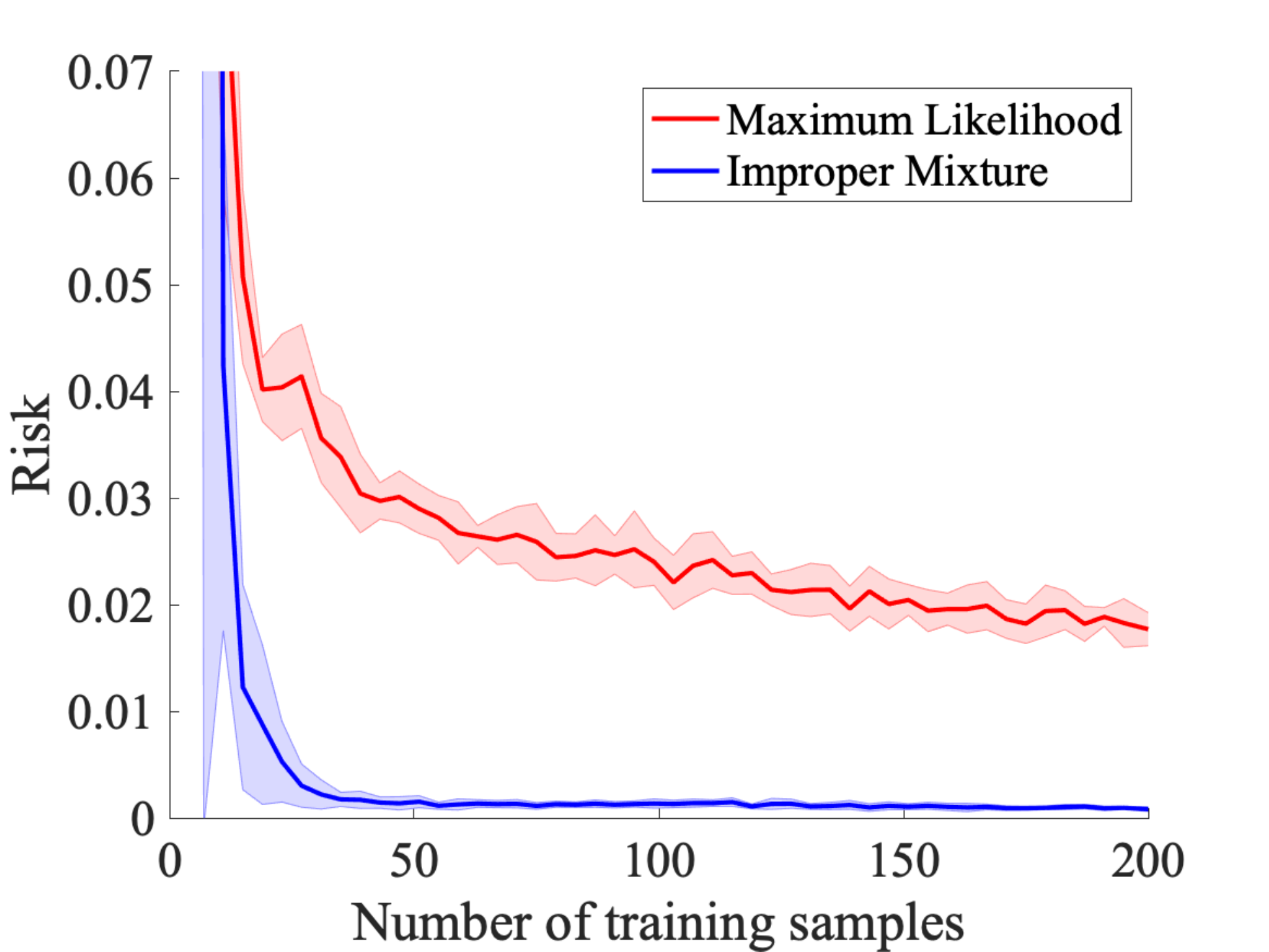} 
\caption{$\tau = 0 \%$ }
\end{subfigure} %
\begin{subfigure}{0.32\textwidth}
  \includegraphics[height=1.7in]{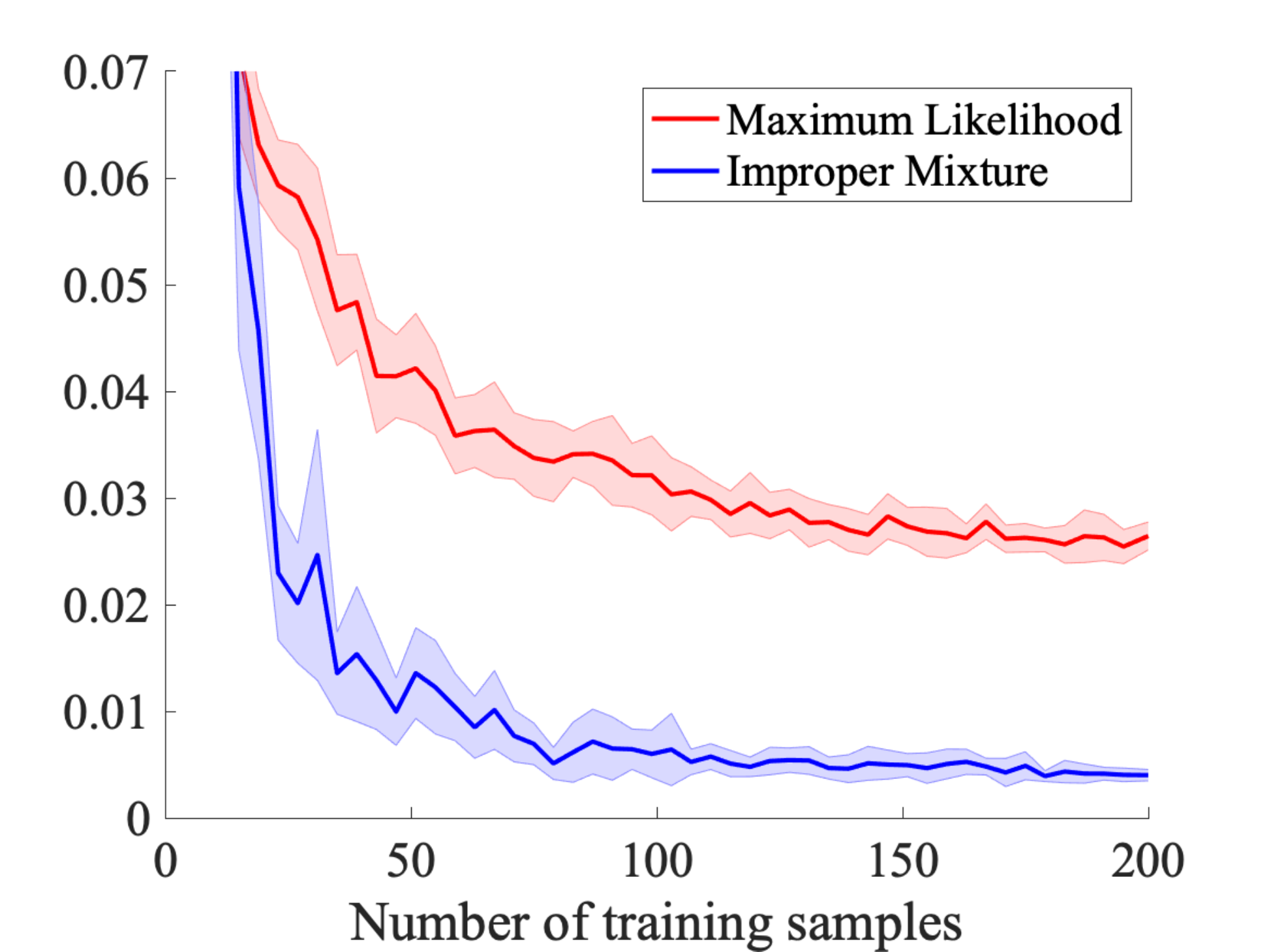} 
\caption{$ \tau = 5\%$ }
\end{subfigure} %
\begin{subfigure}{0.32\textwidth}
  \includegraphics[height=1.7in]{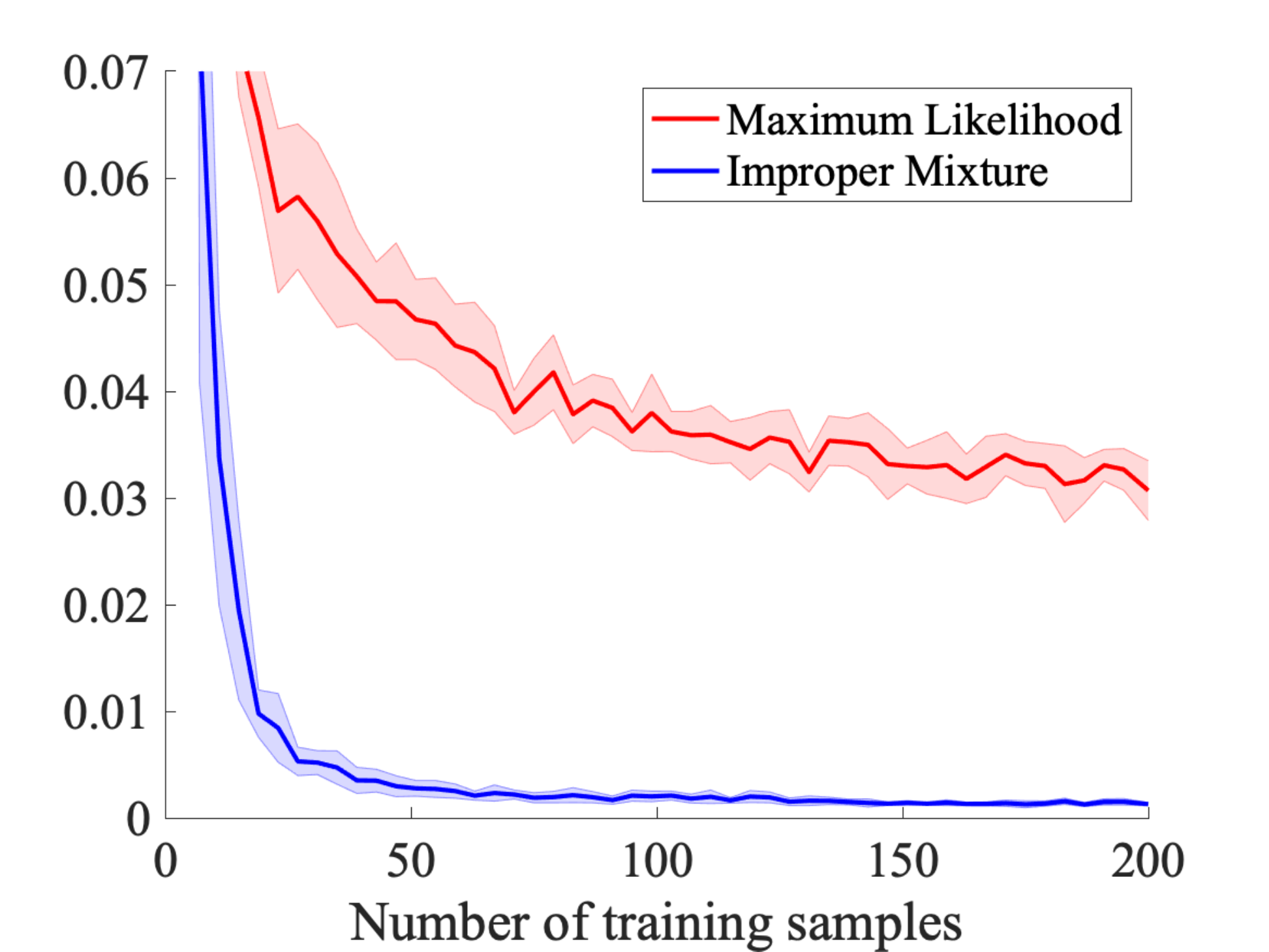} 
\caption{$ \tau = 20\%$ }
\end{subfigure} 
\caption{Logistic Regression, MNIST Data mixed with typed data. As
  misspecification $\tau$ increases, the improper learning algorithm AHA's
  performance remains stable, while the best regularized MLE's performance
  worsens.}
\label{fig2}
\end{figure}

Within each experiment, we
implement several regularization schedules. We test $\paramrad = c$,
$\paramrad = c \log n$, $\paramrad = c \sqrt{n}$, and $\paramrad = c n$ for
$c \in \{ 0.1, 0.2, 1 \}$. In Figures \ref{fig1} and \ref{fig2} \emph{we
  only show the results for the best choice of $\paramrad$ according to the
  performance of the Maximum Likelihood Estimator.} We repeat the experiment
$100$ times on the synthetic data and $10$ times on the real dataset and
average the results. We run the experiment for $\tau = 0, 2.5, 5$ on the
synthetic dataset and $\tau = 0 \%, 5 \%, 20 \%$ for the real dataset.

The results of Figure~\ref{fig1} and Figure~\ref{fig2} are consistent with
our expectations: as the magnitude of misspecification (as measured by
$\tau \ge 0$) increases, the gap in performance between the maximum
likelihood estimator and the aggregated solution increases.  Even more, if
we may be so bold, the results suggest that using a subsampling and
aggregation strategy as in Alg.~\ref{alg:aha} may be a useful primitive for
other learning problems; we leave this as a possibility for future work.


%% file: sections/discussion.tex
\section{Discussion}

This work takes steps toward addressing the fundamental and practically
important challenge of the cost of inaccurate modeling. While
modeling assumptions are ubiquitous throughout statistics, machine
learning, and data science---allowing analyses that demonstrate
fast convergence rates, efficient algorithms, interpretable conclusions---most
such assumptions are (at least) slightly flawed.
This misspecification can have downsides:
in addition to perhaps faulty conclusions from a faulty model, even
convergence rates of estimators may degrade.
This adds a wrinkle to data-modeling tasks: not
only must we choose a model that closely fits the data, but we must be
mindful of the cost of model misspecification, as this cost is not uniform
across models. Our development of the linearity constant $\linearconstfull$
in Eq.~\eqref{eqn:def-linearconst} of the model family gives a reasonably
concise description of potential sensitivity to misspecification for
many model families.

Yet as we additionally consider, for probabilistic prediction problems
aggregation strategies can at least ameliorate these challenges.
Of course, aggregation approaches are familiar throughout statistical
learning~\cite{Tsybakov04, DalalyanTs08, VanErvenGrMeReWi15}, but
we believe their potential for improvement \emph{beyond} ``optimal''
point estimators remains unexplored; our results provide one lens for
viewing this problem.

%% file: sections/proof-lowerbound.tex

\section{Proof of Theorem~\ref{thm:lowerbound}}
\label{append:lowerbound}

Before we give the proof of the theorem proper, we first recall Le Cam's
method. As we consider the excess loss, central to our development
is the following separation quantity~\cite[cf.][Sec.~5]{Duchi18}.
\begin{definition} [Separation]
  \label{def:sep}
  Let $f_{1}: \Theta \to \R$, $f_{2}: \Theta \to \R$. Their
  \emph{separation} with respect to $\Theta$ is
  \begin{equation*}
    \sep(f_{1},f_{2}, \Theta) := \sup \bigg \{ \varepsilon \ge 0
    \mid   \begin{array}{l} f_{1}(\theta)
      \le \inf_{\theta \in \Theta} f_1(\theta) + \varepsilon
      ~~ \mbox{implies} ~~ f_2(\theta) > \inf_{\theta \in \Theta} f_2(\theta)
      + \varepsilon \\
      f_{2}(\theta) \le \inf_{\theta \in\Theta} f_2(\theta)
      + \varepsilon ~~\mbox{implies}~~
      f_{1}(\theta) > \inf_{\theta \in \Theta} f_1(\theta) + \varepsilon,
    \end{array}
    \mbox{all~}\theta \in \Theta
    \bigg\}.
  \end{equation*}
\end{definition}
\noindent
This separation measures the extent to which minimizing a function
$f_1$ means that one cannot minimize a function $f_2$, and
by a standard reduction of estimation and optimization to testing---if one can
optimize well, then one can decide whether one is optimizing $f_1$ or
$f_2$---we have Le Cam's method. (See~\cite[Sec.~5.2]{Duchi18} for this specific
form.)
\begin{lemma} [Le Cam's Method]
  \label{lemma:lecam}
  Let $v \in \{ \pm 1 \}$ and $P_v$
  be arbitrary distributions on a set $\mc{Z}$ and
  $f_v : \Theta \to \R$ be functions similarly indexed by $v \in \{\pm 1\}$,
  where $f_v\opt = \inf_{\theta \in \Theta} f_v(\theta)$.
  Then
  \begin{equation*}
    \inf_{\what{\theta}} \max_{v \in \{-1, 1\}}
    \E_{P_v^n}\left[f_v(\what{\theta}(Z_1, \ldots, Z_n)) - f_v\opt\right]
    \ge \sep(f_1, f_{-1}, \Theta) \left(1 - \sqrt{\frac{n}{2}
      \dkl{P_1}{P_{-1}}}\right),
  \end{equation*}
  where the infimum is over $\what{\theta} : \mc{Z}^n \to \Theta$
  and the expectation is over $Z_i \simiid P_v$.
\end{lemma}

To use Lemma~\ref{lemma:lecam} to prove lower bounds, then, the key is to
show that for a given loss $\loss$, there are distributions $P_1, P_{-1}$
that induce a large separation in the risks $\risk_{P_v}$ while having small
KL-divergence. The basic approach, familiar from other lower
bounds~\cite{Duchi18, Wainwright19}, is to show that for some constants $0 <
c_0, c_1 < \infty$ and a power $\beta \ge 0$, we can choose $P_{\pm 1}$ to
scale with a desired rate $\varepsilon$ via
\begin{equation*}
  \sep(\risk_{P_1}, \risk_{P_{-1}}, \Theta)
  \ge c_0 \varepsilon^\beta
  ~~ \mbox{while} ~~
  \dkl{P_1}{P_{-1}} \le c_1 \varepsilon^2.
\end{equation*}
Given these separation and divergence bounds, it is then evidently the case
that we may choose $\varepsilon^2 = \frac{1}{2 c_1 n}$, which
immediately yields a lower bound via Lemma~\ref{lemma:lecam}
scaling as
\begin{equation*}
  c_0 \left(\frac{1}{2 c_1 n}\right)^{\beta / 2}.
\end{equation*}
Thus any lower bounds we prove become larger as the separation rate $\beta$
decreases or constant $c_0$ grows.
The next lemma does precisely this, though there is some sophistication
required because of the different constraints on our losses.

\begin{lemma}
  \label{lem:linear}
  Let the loss take the form $\loss(p_{\theta}(y \mid x) ) =
  \scalarloss(\theta^{T}x, y)$.
  Let $\varepsilon \in [0, \frac{3}{5}]$, $y \in \mc{Y}$, and $t \in \R$,
  and $\qworst(t, y)$ be as in definition~\eqref{eqn:qworst}.
  Assume $t$ and $\delta \ge 0$ jointly satisfy
  \begin{equation}
    \label{eqn:annoying-conditions-on-delta}
    \sup_{|\Delta| \le \delta}
    \delta \scalarloss''(t + \Delta, y) \le
    \varepsilon \qworst(t,y) |\scalarloss'(t, y)|
    ~~~ \mbox{and} ~~~
    2(t^2 + \delta^2) \le \datarad^2 \paramrad^2.
  \end{equation}
  Then for any $\mc{X} \supset \{x \in\R^d \mid \ltwo{x} \le \datarad\}$,
  there exist
  distributions $\{P_{\pm 1}\}$ on $\mc{X} \times \mc{Y}$
  such that
  \begin{equation*}
    \sep(\risk_{P_1}, \risk_{P_{-1}}, \Theta) \ge
    \frac{\qworst(t, y)}{2} |\scalarloss'(t, y)| \delta \cdot \varepsilon
  \end{equation*}
  while
  \begin{equation*}
    \dkl{P_1}{P_{-1}} \leq \qworst(t, y) \varepsilon^2.
  \end{equation*}
\end{lemma}
\noindent
We prove Lemma \ref{lem:linear} in Appendix \ref{append:linear}.

Now we leverage Lemma~\ref{lem:linear} to provide a minimax
risk bound over $\perturbsize$ variation distance perturbations.
The key here is that the family $\{P_\theta\}$ restricts only
conditional distributions---the marginal distribution over
$X \in \mc{X}$ may be arbitrary---allowing us to give appropriate mixtures.
\begin{lemma}
  \label{lemma:pert}
  Assume that $\mc{X} \supset \{x \in \R^d \mid \ltwo{x} \le \datarad\}$ and
  let $P_{\pm 1}$ be distributions on $\mc{X} \times \mc{Y}$.
  Let
  $\perturbsize \in [0, 1]$.  Then there exists a distribution $P_0 \in
  \{P_\theta\}_{\theta \in \Theta}$ such that for $Q_{\pm \perturbsize}
  \defeq (1 - \perturbsize) P_0 + \perturbsize P_{\pm 1}$,
  \begin{equation*}
    \sep(\risk_{Q_\perturbsize}, \risk_{Q_{-\perturbsize}}, \Theta)
    = \perturbsize \sep(\risk_{P_1}, \risk_{P_{-1}}, \Theta)
    ~~ \mbox{and} ~~
    \dkl{Q_\perturbsize}{Q_{-\perturbsize}} \le \perturbsize
    \dkl{P_1}{P_{-1}}.
  \end{equation*}
\end{lemma}
\noindent
See Appendix~\ref{append:pert} for the short proof of the result.

With Lemma~\ref{lemma:pert} in hand we can now prove Theorem
\ref{thm:lowerbound}.

\subsection{Proof of Theorem~\ref{thm:lowerbound} proper}

First, recalling the perturbed minimax risk
from Definition \ref{def:minimax-risk},
\begin{align*}
  \minimax_n(\Theta, \playdists, \perturbsize)
  \defeq  \inf_{\what{p}_n \in \Gamma}
  \sup_{\theta \in \Theta}
  \sup_{P : \tvnorms{P - P_\theta} \le \perturbsize} 
  \E_{P^n}[\risknorm_P(\what{p}_n)],
\end{align*}
where the infimum is over all procedures.  Now, let $(\varepsilon, y, t,
\delta)$ be any collection satisfying the conditions of
Lemma~\ref{lem:linear} and $\{P_{\pm 1}\}$ be the distributions the lemma
guarantees exist.  Additionally, let $Q_{\pm \perturbsize}$ be the perturbed
distributions Lemma~\ref{lemma:pert} provides, so that there exists $P_0 \in
\{P_\theta\}$ such that $\tvnorm{P_0 - Q_{\pm \perturbsize}} \le
\perturbsize \tvnorm{P_0 - P_{\pm 1}} \le \perturbsize$.
Then we immediately obtain
\begin{align*}
  \minimax_n(\Theta, \playdists, \perturbsize)
  & \ge \inf_{\what{\theta}_n}
  \max_{v \in \pm 1}
  \E_{Q^n_{v \perturbsize}} \left[\risknorm_{Q_{v \perturbsize}}(\what{\theta}_n)\right] \\
  & \stackrel{(i)}{\ge}
  \sep(\risk_{Q_\perturbsize}, \risk_{Q_{-\perturbsize}}, \Theta)
  \left(1 - \sqrt{\frac{n}{2} \dkl{Q_\perturbsize}{Q_{-\perturbsize}}}\right) \\
  & \stackrel{(ii)}{\ge} \perturbsize
  \sep(\risk_{P_1}, \risk_{P_{-1}}, \Theta)
  \left(1 - \sqrt{\frac{n \perturbsize}{2} \dkl{P_1}{P_{-1}}}\right) \\
  & \stackrel{(iii)}{\ge}
  \frac{\perturbsize \qworst(t, y) |\scalarloss'(t, y)| \delta}{2}
  \varepsilon \left(1 - \sqrt{n \perturbsize \qworst(t, y) \varepsilon^2 / 2}
  \right),
\end{align*}
where inequality~$(i)$ is Le Cam's inequality (Lemma~\ref{lemma:lecam}),
inequality~$(ii)$ follows via Lemma~\ref{lemma:pert}, and
Lemma~\ref{lem:linear} gives inequality~$(iii)$ whenever $\varepsilon \le
\frac{3}{5}$. Choosing $\varepsilon^2 = \frac{1}{2 n \perturbsize
  \qworst(t, y)}$ (where we use that $n$ is large enough that
$\varepsilon^2 \le \frac{1}{3}$) yields
the lower bound
\begin{equation}
  \label{eqn:intermediate-minimax}
  \minimax_n(\Theta, \Gamma, \perturbsize)
  \ge \frac{\sqrt{\perturbsize \qworst(t, y)}}{4 \sqrt{n}}
  |\scalarloss'(t, y)| \delta
\end{equation}
valid for all $\delta \ge 0$ satisfying
\begin{equation*}
  \delta \le \frac{|\scalarloss'(t, y)|}{\sup_{|\Delta| \le \delta}
    \scalarloss''(t + \Delta, y)}
  \frac{\sqrt{\qworst(t, y)}}{\sqrt{2 n \perturbsize}}.
\end{equation*}
This is circular, but we note
that if we define
\begin{equation*}
  m_n(\delta)
  = m_n(\delta, t, y, \scalarloss, \perturbsize)
  \defeq \min\left\{\delta, \frac{|\scalarloss'(t, y)|}{
    \sup_{|\Delta| \le \delta}
    \scalarloss''(t + \Delta, y)}
  \frac{\sqrt{\qworst(t, y)}}{\sqrt{2 n \perturbsize}}\right\}
\end{equation*}
then $m_n(\delta)$ satisfies
$m_n(\delta) \le \frac{|\scalarloss'(t, y)|}{\sup_{|\Delta| \le
    m_n(\delta)} \scalarloss''(t + \Delta, y)} \frac{\sqrt{\qworst(t,
    y)}}{\sqrt{2 n \perturbsize}}$, and substituting $m_n(\delta)$ for
$\delta$ in the lower
bound~\eqref{eqn:intermediate-minimax} gives
\begin{equation*}
  \minimax_n(\Theta, \playdists, \perturbsize)
  \ge \frac{\sqrt{\perturbsize \qworst(t, y)}}{4 \sqrt{n}}
  |\scalarloss'(t, y)|
  \min\left\{\delta, \frac{|\scalarloss'(t, y)|}{
    \sup_{|\Delta| \le \delta}
    \scalarloss''(t + \Delta, y)}
  \frac{\sqrt{\qworst(t, y)}}{\sqrt{2 n \perturbsize}}\right\},
\end{equation*}
valid for all $\delta \ge 0$ satisfying
$2(t^2 + \delta^2) \le \datarad^2 \paramrad^2$
as in Eq.~\eqref{eqn:annoying-conditions-on-delta}.

%% file: sections/appendix-linearlemma.tex

\subsection{Proof of Lemma \ref{lem:linear}}
\label{append:linear}

Recall throughout that $\varepsilon \in [0, 1]$.
We provide the proof in two parts. In the first, we demonstrate
the claimed risk separation by a Taylor approximation argument, and in the
second, we provide the claimed bound on the KL divergence.

To show the risk separation, choose orthogonal vectors $v, w \in \R^d$
satisfying $\ltwo{v} = \ltwo{w} = \datarad / \sqrt{2}$ and $\<v, w\> = 0$,
so that $\ltwo{v \pm w} = \datarad$. For values $q \in [0, 1]$,
$\alpha \in [-1, 1]$, and $y_0 \in
\mc{Y}$ to be specified presently, we consider distributions on $\R^d \times
\mc{Y}$ defined for $\sigma \in \{-1, 0, 1\}$ by
\begin{equation}
  \label{eqn:pi-distributions}
  P_i :
  (X, Y) = \begin{cases} (\alpha v, y_0) & \mbox{with~probability~} 1 - q \\
    (v + w, y) & \mbox{with~probability~} \frac{q}{2} (1 + \sigma \varepsilon) \\
    (v - w, y) & \mbox{with~probability~} \frac{q}{2} (1 - \sigma \varepsilon).
  \end{cases}
\end{equation}
In this case, the risk evidently satisfies
\begin{equation*}
  \risk_{P_0}(\theta) = (1 - q) \scalarloss(\alpha \theta^T v, y_0)
  + \frac{q}{2} \left[\scalarloss(\theta^T(v + w), y)
    + \scalarloss(\theta^T (v - w), y)\right].
\end{equation*}
We now construct its minimizer by judicious choice of $q$, where
scaling by $\alpha \in [-1, 1]$ is sometimes necessary. Define
$\theta_0 = \frac{2}{\datarad^2} t v$, so that
$\ltwo{\theta_0} = \sqrt{2} t / \datarad \le \paramrad$,
$\theta_0^T w = 0$ and $\theta_0^T v = t$, and
\begin{equation*}
  \nabla \risk_{P_0}(\theta_0) =
  \alpha (1 - q) \scalarloss'(\alpha t, y_0) v + q \scalarloss'(t, y) v,
\end{equation*}
so that if
\begin{equation*}
  q = \frac{\alpha \scalarloss'(\alpha t, y_0)}{
    \alpha \scalarloss'(\alpha t, y_0) - \scalarloss'(t, y)}
\end{equation*}
satisfies $q \in [0, 1]$,
we have $\nabla \risk_{P_0}(\theta_0) = 0$ and $\theta_0
\in \argmin_{\theta \in \Theta} \risk_{P_0}(\theta)$. In particular, we may choose
\begin{equation*}
  q = \qworst(t,y) \defeq \sup_{y_0 \in \mc{Y}, \alpha \in [-1, 1]}
  \left\{\frac{\alpha \scalarloss'(\alpha t, y_0)}{
    \alpha \scalarloss'(\alpha t, y_0) - \scalarloss'(t, y)}
  ~ \mbox{s.t.}~
  \sign(\alpha \scalarloss'(\alpha t, y_0))
  \neq \sign(\scalarloss'(t, y)) \right\}.
\end{equation*}

We will perform a Taylor approximation of the risks $\risk_{P_{\sigma}}$ for $\sigma \in \{ \pm 1 \}$ around
$\theta_0$ to show the desired separation bound.
To that end, for $\delta \in \R$ define the shifted vector
\begin{equation*}
  \theta_\delta \defeq \frac{2}{\datarad^2} (t v + \delta w)
  = \theta_0 + \frac{2 \delta}{\datarad^2} w,
\end{equation*}
for which we have $\ltwo{\theta_\delta}^2 = 2(t^2 + \delta^2) / \datarad^2$
and $\theta_\delta^T(v \pm w) = t \pm \delta$.
Using the risk expansion
\begin{equation}
  \label{eqn:risk-expansion}
  \risk_{P_{\sigma}}(\theta)
  = \risk_{P_0}(\theta)
  + \frac{q  \sigma \varepsilon}{2}
  \left[\scalarloss(\theta^T (v + w), y) - \scalarloss(\theta^T(v - w), y)
    \right]
\end{equation}
and the Taylor approximation
\begin{equation*}
  \scalarloss(t + \delta, y)
  = \scalarloss(t, y) + \scalarloss'(t, y) \delta
  + \frac{\delta^2}{2} \scalarloss''(t + \Delta, y)
  ~~ \mbox{for~some~}\Delta \in [0, \delta],
\end{equation*}
we obtain
\begin{align*}
    \risk_{P_{\sigma}}(\theta_\delta)
    & = (1 - q) \scalarloss(t, y_0) + q \scalarloss(t, y)
    + \sigma \varepsilon q \scalarloss'(t, y) \cdot \delta
    + \frac{\delta^2}{2} \mbox{rem}(\delta) \\
    & = \risk_{P_0}(\theta_0)
    + \sigma \varepsilon q \scalarloss'(t, y) \cdot \delta
    + \frac{\delta^2}{2} \mbox{rem}(\delta),
\end{align*}
where the remainder term $|\mbox{rem}(\delta)| \le
\sup_{|\Delta| \le \delta} \scalarloss''(t + \Delta, y)$.
In particular, if $|\delta|$ is small enough that
the conditions~\eqref{eqn:annoying-conditions-on-delta} hold,
that is,
\begin{equation*}
  \sup_{|\Delta| \le |\delta|}
  |\delta| \scalarloss''(t + \Delta, y) \le
  \varepsilon q |\scalarloss'(t, y)|
  ~~~ \mbox{and} ~~~
  2(t^2 + \delta^2) \le \datarad^2 \paramrad^2,
\end{equation*}
then setting $s = -\sign(\sigma \scalarloss'(t, y))$ and letting $\delta \ge 0$
satisfy the conditions~\eqref{eqn:annoying-conditions-on-delta}, we have
$\theta_{s\delta} \in \Theta$ and
\begin{equation*}
  \inf_{\theta \in \Theta}
  \risk_{P_{\sigma}}(\theta)
  \le \risk_{P_{\sigma}}(\theta_{s \delta})
  \le \risk_{P_0}(\theta_0) - \frac{q \varepsilon}{2} |\scalarloss'(t, y)| \delta.
\end{equation*}
Combining this inequality with the risk
expansion~\eqref{eqn:risk-expansion}, we see immediately that if
$\theta \in \Theta$ satisfies $\scalarloss(\theta^T(v + w), y) \ge
\scalarloss(\theta^T(v - w), y)$
then
\begin{equation*}
  \risk_{P_1}(\theta) \ge
  \inf_{\theta \in \Theta} \risk_{P_1}(\theta) + \frac{q \varepsilon}{2}
  |\scalarloss'(t, y)| \delta,
\end{equation*}
and conversely
$\scalarloss(\theta^T(v - w), y) \le
\scalarloss(\theta^T(v - w), y)$ implies
\begin{equation*}
  \risk_{P_{-1}}(\theta) \ge
  \inf_{\theta \in \Theta} \risk_{P_1}(\theta) + \frac{q \varepsilon}{2}
  |\scalarloss'(t, y)| \delta.
\end{equation*}

As $\theta_0$ minimizes $\risk_{P_0}$, the expansion~\eqref{eqn:risk-expansion}
implies that any $\theta$ minimizing
$\risk_{P_i}(\theta)$ over $\Theta$ necessarily satisfies
$\sigma [\scalarloss(\theta^T(v + w), y) - \scalarloss(\theta^T(v - w), y)]
< 0$, so we obtain the risk separation
\begin{equation*}
  \sep(\risknorm_{P_1}, \risknorm_{P_{-1}}, \Theta)
  \ge \frac{q \epsilon}{2} |\scalarloss'(t, y)| \delta ,
\end{equation*}
valid for any $\delta$ satisfying the
constraints~\eqref{eqn:annoying-conditions-on-delta}, which proves
the claimed risk separation in the lemma.

To see the KL bound in Lemma~\ref{lem:linear}, we note that for any pair of
distributions of the form~\eqref{eqn:pi-distributions},
we have
\begin{equation*}
  \dkl{P_1}{P_{-1}}
  = \frac{q(1 + \varepsilon)}{2}
  \log \frac{1 + \varepsilon}{1 - \varepsilon}
  + \frac{q(1 - \varepsilon)}{2}
  \log \frac{1 - \varepsilon}{1 + \varepsilon}
  = q \varepsilon \log \frac{1 + \varepsilon}{1 - \varepsilon}
  \stackrel{(\star)}{\le} q \varepsilon^2,
\end{equation*}
where inequality~$(\star)$ is valid for $\varepsilon \le \frac{3}{5}$.

%% file: sections/proof-perturbation-separation.tex
\subsection{Proof of Lemma~\ref{lemma:pert}}
\label{append:pert}

Let $P_0$ have any distribution on $Y \mid X$ and $P_0(X = \zeros) = 1$,
that is, the marginal over $X$ is supported completely on $\zeros$.
Then it is immediate that
for $Q_{\pm \perturbsize} = (1 - \perturbsize) P_0 + \perturbsize P_{\pm 1}$,
we have
\begin{equation*}
  \risk_{Q_{\pm \perturbsize}}(\theta)
  = (1 - \perturbsize) \E_{P_0}[\scalarloss(0, Y)]
  + \perturbsize \risk_{P_{\pm 1}}(\theta),
\end{equation*}
and therefore $\risknorm_{Q_{\pm \perturbsize}}(\theta) = \perturbsize
\risknorm_{P_{\pm 1}}(\theta)$.
It is therefore immediate that
$\sep(\risk_{Q_\perturbsize}, \risk_{Q_{-\perturbsize}}, \Theta)
= \perturbsize \sep(\risk_{P_1}, \risk_{P_{-1}}, \Theta)$.
For the gap on the KL divergence, we use joint convexity
to obtain
\begin{equation*}
  \dkl{Q_\perturbsize}{Q_{-\perturbsize}}
  = \dkl{(1 - \perturbsize) P_0 + \perturbsize P_1}{
    (1 - \perturbsize) P_0 + \perturbsize P_{-1}}
  \le (1 - \perturbsize) \underbrace{\dkl{P_0}{P_0}}_{= 0}
  + \perturbsize \dkl{P_1}{P_{-1}}.
\end{equation*}

%% file: sections/proof-slow-rates-oops.tex
\subsection{Proof of Proposition~\ref{proposition:slow-rates-oops}}
\label{sec:proof-slow-rates-oops}

By assumption, there exist
$t, y, y_0$ satisfying $\scalarloss'(t, y) \scalarloss'(t, y_0) < 0$,
and $\scalarloss''(t, y) = 0$. Then it is evidently the case that
$\qworst \equiv \qworst(t, y) > 0$, so
that we obtain the lower bound
\begin{equation*}
  \linearconstfull
  \ge c \sup_{0 \le \delta \le \datarad \paramrad / 2}
  |\scalarloss'(t, y)|
  \min\left\{\delta \sqrt{\perturbsize \qworst},
  \frac{|\scalarloss'(t, y)|}{
    \sup_{|\Delta| \le \delta} \scalarloss''(t + \Delta, y)}
  \frac{\qworst}{\sqrt{2n}}\right\}.
\end{equation*}
Now, we recall that $\scalarloss$ is $\mc{C}^3$ near
$t$ and by assumption $\scalarloss''(t, y) = 0$,
for all suitably small $\delta$ we obtain
$|\scalarloss''(t + \Delta, y)| \ge |\scalarloss'''(t, y)| |\Delta|/2$,
and so in particular for all small $\delta$,
\begin{equation*}
  \linearconstfull
  \ge c 
  |\scalarloss'(t, y)|
  \min\left\{\delta \sqrt{\perturbsize \qworst},
  \frac{2 |\scalarloss'(t, y)|}{|\scalarloss'''(t, y)| \delta}
  \frac{\qworst}{\sqrt{2n}}\right\}.
\end{equation*}
Set $\delta^2 = \frac{1}{\sqrt{n}}$ to obtain that
for some problem-dependent constant $c_{\textup{prob}}$, we have
$\linearconstfull \ge c_{\textup{prob}}\frac{1}{n^{1/4}}$.
Substitute this lower bound in Theorem~\ref{thm:lowerbound}.

%% file: sections/appendix-mixconstproofs.tex

\subsection{Proofs of mixability in Table~\ref{table:examples}}
\label{append:mix}

\newcommand{\simplex}{\Delta}
\providecommand{\ones}{\mathbf{1}}

We assume that $\mc{Y}$ is discrete and of size $k$ (it is not difficult to
obtain a result when $\mc{Y} = \N$), so that we may identify distributions on
$\mc{Y}$ with vectors $p \in \simplex_k \defeq \{v \in R^k_+ \mid \ones^T v =
1\}$, the probability simplex in $\R^k$.  Consider any $\mc{C}^2$ function $h
: \simplex \to \R$, noting that
\begin{align*}
  \nabla \exp(-\eta h(p))
  & = - \eta \exp(-\eta h(p)) \nabla h(p), \\
  \nabla^2 \exp(-\eta h(p))
  & = \eta \exp(-\eta h(p))
  \left[\eta \nabla h(p) \nabla h(p)^T - \nabla^2 h(p)\right].
\end{align*}
We consider each of the columns of the table in turn. Thus to demonstrate
exp-concavity it is sufficient that $\nabla^2 h(p) \succeq \eta \nabla h(p)
\nabla h(p)^T$ for all $p \in \simplex_k$.
\begin{enumerate}[1.]
\item For $\logloss$, we take $h(p) = -\log p$, for which it is immediate
  that $\eta = 1$ suffices as $\exp(-h(p)) = p$.
\item For $\losssq$, we have $h(p) = \half (p - 1)^2$,
  $h'(p) = (p - 1)$, and $h''(p) = 1$, so $\eta = 1$
  suffices.
\item For $\losshel$, we have
  $h(p) = (\sqrt{p} - 1)^2
  = p - 2 \sqrt{p} + 1$,
  $h'(p) = 1 - \frac{1}{\sqrt{p}}$, and
  $h''(p) = \frac{1}{2 p^{3/2}}$. Thus, we seek
  $\eta$ such that
  \begin{equation*}
    \frac{1}{2 p^{3/2}} \ge \eta (1 - 1/\sqrt{p})^2
    ~~ \mbox{or} ~~
    \half \ge \eta (p^{3/2} - 2p + \sqrt{p})
  \end{equation*}
  for all $p \in [0, 1]$. Letting $\beta = \sqrt{p}$ and solving for
  the stationary points of $\beta^3 - 2\beta + \beta$
  at $\sqrt{p} = \beta = 1/3$ and $\beta = 1$,
  we see it is sufficient that $1 \ge 2 \eta (1/27 - 2/9 + 1/3)
  = \frac{8}{27} \eta$, or $\eta \le \frac{27}{8}$.
\item For $\lossquad$, we have $h(p) = \half \ltwo{p - e_y}^2$,
  so it suffices that $I - \eta (p - e_y) (p - e_y)^T \succeq 0$, or
  $\eta \le \half$.
\end{enumerate}


%% file: sections/appendix-optseqglm2.tex
\subsection{Proof of Theorem~\ref{theorem:well-specified-lower}}
\label{appendix:optseq}


Recall Definition~\ref{def:sep} of the separation between two functions.  We
first recall the essentially standard reduction of estimation to testing,
which proceeds as follows. Let $\mc{V}$ be a finite set indexing a
collection $\{P_v\}_{v \in \mc{V}}$ of distributions on $\mc{X} \times
\mc{Y}$ and a collection of functions $\{f_v\}$.  Consider the following
process: draw $V \in \mc{V}$ uniformly at random, and conditional on $V =
v$, observe $(X_i, Y_i) \simiid P_v$ for $i = 1, 2, \ldots, n$.  Then we
have the following lemma, which reduces optimization of $f_v$ to testing
the index $V$ (see, e.g.~\cite[Sec.~5]{Duchi18} or
\cite[Ch.~15]{Wainwright19}).
\begin{lemma}
  \label{lemma:optimization-to-testing}
  Let $f_v\opt = \inf_{\theta \in \Theta} f_v(\theta)$ for $v \in \mc{V}$. Then
  \begin{equation*}
    \inf_{\what{\theta}_n}
    \max_{v \in \mc{V}} \E_{P_v^n}\left[f_v(\what{\theta}_n(X_1^n, Y_1^n))
      - f_v\opt\right]
    \ge \min_{v \neq w \in \mc{V}} \sep(f_v, f_w, \Theta)
    \cdot \inf_{\what{\Psi}_n} \P(\what{\Psi}_n(X_1^n, Y_1^n) \neq V),
  \end{equation*}
  where the infima are over procedures $\what{\theta}_n : \mc{X}^n \times
  \mc{Y}^n \to \Theta$ and all measurable functions $\what{\Psi}_n$,
  respectively.
\end{lemma}

\noindent
We thus lower bound the probability of error in testing,
$\what{\Psi} \neq V$, for which we use Fano's
inequality~\cite{CoverTh06}:
\begin{lemma}[Fano's Inequality]
  \label{lemma:fano}
  Let $I(V; X_{1}^{n},Y_{1}^{n})$ be the (Shannon) mutual information
  between $V$ and $(X_{1}^{n}, Y_{1}^{n})$, where $(X_i, Y_i) \simiid P_v$
  conditional on $V = v$ and $V$ is uniform on $\mc{V}$. Then
  for any $\what{\Psi}$,
  \begin{equation*}
    \P(\what{\Psi}(X_1^n, Y_1^n) \neq V) \ge
    1 - \frac{I(V; X_{1}^{n},Y_{1}^{n}) + \log 2}{\log |\mc{V}| }.
  \end{equation*}
\end{lemma} 

Now, we define the collection of problems we consider and their induced
risks. Let $X$ be uniform on $\{\pm 1\}^d$, and let
\begin{equation*}
  p_\theta(y \mid x) = \exp(y \theta^T x - A(\theta^T x))
\end{equation*}
be the density of $P_\theta$ with respect to the base measure $\nu$.
For a value
$\delta \ge 0$ to be chosen, let $P_v$ be the joint distribution
on $(X, Y)$ with $\theta = \delta v$.
We first demonstrate that these induce a separation in the
expected log loss of a predictive distribution $p(\cdot \mid x)$,
where for such a $p$ we define
the risk
\begin{equation*}
  \risk_{\delta v}(p) \defeq \E_{P_v}\left[\logloss(p(\cdot \mid X), Y)\right]
  = \E_{P_v}[-\log p(Y \mid X)],
\end{equation*}
where we note that $p_{\delta v}$ minimizes $\risk_{\delta v}$ as it is
well-specified.  The key to applying
Lemmas~\ref{lemma:optimization-to-testing} and~\ref{lemma:fano}
are the following two technical results, which
respectively lower bound the separation and upper bound
the KL-divergence between distributions.
We defer proofs to Sections~\ref{sec:proof-separation}
and~\ref{sec:proof-kl-bound}.

\begin{lemma}
  \label{lemma:separation}
  Let $\mc{P}$ be the collection of all
  conditional probability distributions
  on $Y \mid X$. There exists a constant $C(A)$ depending only
  on the log partition function $A(\cdot)$
  such that for all $\delta \ge 0$ and $u, v \in \R^d$,
  \begin{equation*}
    \sep(\risk_{\delta v}, \risk_{\delta w}, \mc{P})
    \ge \frac{1}{16} A''(0) \delta^2 \ltwo{v - w}^2
    - C(A) \delta^3 d^{3/2} \max\{\ltwo{v}, \ltwo{w}\}^3.
  \end{equation*}
\end{lemma}

\begin{lemma}
  \label{lemma:kl-bound}
  For $v \in \R^d$, let $P_{\delta v}$ denote
  the joint distribution over $X \sim \uniform(\{-1, 1\}^d)$ and
  $Y \mid X = x$ having exponential family
  density $p_{\delta v}(y \mid x) = \exp(y \theta^T x - A(\theta^T x))$.
  There exists a constant $C(A)$ depending only on the log partition
  function $A(\cdot)$ such that for all $\delta \in [0, 1]$ and $u, v$
  satisfying $\ltwo{u} \le 1$, $\ltwo{v} \le 1$,
  \begin{equation*}
    \dkl{P_{\delta v}}{P_{\delta w}} \le \frac{\delta^2}{2}
    A''(0) \ltwo{v - w}^2 + C(A) \delta^3 \ltwo{v - w}^3.
  \end{equation*}
\end{lemma}

With these two lemmas, the result is relatively straightforward.
We consider two cases: that $d \ge 8$ and (for completeness) that
$d \le 8$, which we defer temporarily.
Let $d \ge 8$.
By a standard volume argument~\cite[Ch.~15]{Wainwright19},
there exists a packing set $\mc{V} \subset \{v \in \R^d \mid \ltwo{v} = 1\}$
of the $\ell_2$ sphere satisfying
$|\mc{V}| \ge \exp(d / 4)$ and $\ltwo{v - w} \ge \half$
for each $v \neq w \in \mc{V}$. Let $V$ be uniform on $\mc{V}$ as in our
construction above. Then naive bounds on the mutual information
$I(V; X_1^n, Y_1^n)$ yield that
\begin{equation*}
  I(V; X_1^n, Y_1^n)
  \le \frac{1}{|\mc{V}|^2}
  \sum_{v, w \in \mc{V}} \dkl{P_v^n}{P_w^n}
  \stackrel{(\star)}{\le} n \cdot \frac{1}{|\mc{V}|^2}
  \sum_{v, w \in \mc{V}}
  \delta^2 A''(0) \ltwo{v - w}^2
  \le 4 n \delta^2 A''(0),
\end{equation*}
where inequality~$(\star)$ holds for any sufficiently small $\delta \ge 0$
by Lemma~\ref{lemma:kl-bound}.  Applying
Lemmas~\ref{lemma:optimization-to-testing} and \ref{lemma:separation}
by noting that $\ltwo{v - w} \ge \half$, there
exists a numerical constant $c > 0$ such that for small enough $\delta \ge
0$,
\begin{align*}
  \minimax_n(\Theta, \mc{P}, 0)
  & \ge c A''(0) \delta^2
  \inf_{\what{\Psi}_n} \P(\what{\Psi}_n(X_1^n, Y_1^n) \neq V) \\
  & \ge c A''(0) \delta^2 \left(1 - \frac{I(V; X_1^n, Y_1^n) + \log 2}{
    \log |\mc{V}|}\right),
\end{align*}
where the second inequality is Fano's inequality (Lemma~\ref{lemma:fano}).
Applying the preceding bound on the mutual information and that
$\log |\mc{V}| \ge d/4$ then implies
\begin{equation*}
  \minimax_n(\Theta, \mc{P}, 0)
  \ge c A''(0) \delta^2
  \left(1 - \frac{16 n \delta^2 A''(0) + 4 \log 2}{d} \right).
\end{equation*}
Choosing $\delta^2 = \frac{d}{32 A''(0) n}$ then gives the theorem
in the case that $d \ge 8$.

For the final case that $d \le 8$, we apply Le Cam's method
as in our proof of Theorem~\ref{thm:lowerbound}. We assume that $d = 1$,
as increasing the dimension simply increases the risk bound, and
let $X \sim \uniform(\{-1, 1\})$. Recalling
Lemma~\ref{lemma:lecam}, we apply Lemmas~\ref{lemma:separation}
and \ref{lemma:kl-bound} to obtain
\begin{equation*}
  \minimax_n(\Theta, \mc{P}, 0)
  \ge c A''(0) \delta^2 \left(1 - \sqrt{C n \delta^2 A''(0)}\right),
\end{equation*}
where $0 < c$ and $C < \infty$ are numerical constants.
Setting $\delta^2 = \frac{1}{4 C n A''(0)}$ then yields the result.

\subsubsection{Proof of Lemma~\ref{lemma:separation}}
\label{sec:proof-separation}

We define the excess risk functional
\begin{equation*}
  f_{\delta v}(p) \defeq \risk_{\delta v}(p) - \inf_p \risk_{\delta v}(p)
  = \E_{P_v}\left[\log \frac{p_{\delta v}(Y \mid X)}{p(Y \mid X)}\right]
  = \E\left[\dkl{p_{\delta v}(\cdot \mid X)}{p(\cdot \mid X)}\right],
\end{equation*}
where we have used that the exponential
family model $p_{\delta v}$ minimizes $\risk_{\delta v}$,
and we note that
\begin{equation*}
  \sep(f_{\delta v}, f_{\delta w}, \mc{P})
  \ge \half \inf_{p \in \mc{P}}
  \{f_{\delta v}(p) + f_{\delta w}(p)\}
\end{equation*}
(this inequality is valid for any functions and set $\mc{P}$).  
Thus
\begin{equation*}
  2 \sep(\risk_{\delta v}, \risk_{\delta w}, \mc{P})
  = 2 \sep(f_{\delta v}, f_{\delta w}, \mc{P})
  \ge \E\left[\inf_p \left\{\dkl{p_{\delta v}(\cdot \mid X)}{p}
    + \dkl{p_{\delta w}(\cdot \mid X)}{p} \right\}\right].
\end{equation*}
Now we use that for any three distributions
$P_0, P_1, Q$, if $\wb{P} = \half (P_0 + P_1)$ then
\begin{equation*}
  \dkl{P_0}{Q} + \dkl{P_1}{Q}
  = \dkl{P_0}{\wb{P}} + \dkl{P_1}{\wb{P}} + 2 \dkl{\wb{P}}{Q}
  \ge \dkl{P_0}{\wb{P}} + \dkl{P_1}{\wb{P}},
\end{equation*}
and substituting this into the preceding lower bound on the separation gives
\begin{equation}
  \label{eqn:separation-lower-bound}
  \begin{split}
    2 \sep(\risk_{\delta v}, \risk_{\delta w}, \mc{P})
    & \ge \E\left[\dkl{p_{\delta v}(\cdot \mid X)}{
        (1/2) (p_{\delta v}(\cdot \mid X) + p_{\delta w}(\cdot \mid X))}\right] \\
    & \qquad ~ + \E\left[\dkl{p_{\delta w}(\cdot \mid X)}{
        (1/2) (p_{\delta v}(\cdot \mid X) + p_{\delta w}(\cdot \mid X))}
      \right],
  \end{split}
\end{equation}
where the outer expectation is over $X \sim \uniform(\{-1, 1\}^d)$.

We now provide an asymptotic lower bound on the KL divergences, focusing
on a single term given $X = x$ in the lower
bound~\eqref{eqn:separation-lower-bound}. By a Taylor
expansion,
\begin{equation*}
  \log(1 + e^t) = \log 2 + \frac{t}{2} + \frac{t^2}{8} \pm O(1) t^3,
\end{equation*}
where $O(1)$ denotes a universal numerical constant and the expansion is
valid for all $t \in \R$ because $t \mapsto \log(1 + e^t)$ is $1$-Lipschitz.
Using the shorthand $t = \delta v^T x$ and $u = \delta w^T x$
and $p_t(y) = p_{\delta v}(\cdot \mid x)$ and similarly for $p_u$,
we have
\begin{align*}
  \lefteqn{\dkl{p_t}{(1/2) (p_t + p_u)}
    = \int p_t(y) \log \frac{2}{1 + p_u(y) / p_t(y)} d\nu } \\
  & = \int p_t(y) \left[\log 2 - \log\left(1 + e^{y (u - t)
      - (A(u) - A(t))}\right)\right] d\nu \\
  & = \int p_t(y) \left[
    \frac{y(t - u) + A(u) - A(t)}{2}
    - \frac{(y(t - u) + A(u) - A(t))^2}{8}
    \pm O(1) (y(t - u) + A(u) - A(t))^3 \right] d\nu.
\end{align*}
By standard properties of
exponential families, if $\E_t$ denotes expectation under $p_t$,
we have $A'(t) = \E_t[Y]$, and $A$ is $\mc{C}^\infty$ near 0, so that
$A(u) - A(t) = A'(t)(u - t) + \half (u - t)^2 A''(\wt{u})$ for
some $\wt{u} \in [u, t]$.
We may thus write
\begin{align*}
  \lefteqn{\dkl{p_t}{(1/2) (p_t + p_u)}} \\
  & = \int p_t(y) \bigg[\frac{(y - A'(t))(t - u)}{2}
    + \frac{(u - t)^2 A''(\wt{u})}{4}
    - \frac{\left((y - A'(t)) (t - u) + (u - t)^2 A''(\wt{u}) / 2\right)^2}{8}
    \\
    & \qquad \qquad ~~~
    \pm O(1) \left[|y - A'(t)|^3 |t - u|^3
      + (t - u)^6 A''(\wt{u})^3 \right]
    \bigg] d\nu \\
  & = \frac{1}{4} A''(\wt{u}) (u - t)^2
  - \frac{1}{8} A''(t) (u - t)^2
  - \frac{1}{32} A''(\wt{u})^2 (u - t)^4
  \pm O(1) \E_t[|Y - \E_t[Y]|^3] |t - u|^3.
\end{align*}
As $A(\cdot)$ exists in a neighborhood of 0, the moment generating
functions of $p_t, p_u$ exist, this expansion is uniform
in $u, t$ near 0, and so we obtain
\begin{equation}
  \label{eqn:kl-expansions-fun}
  \dkl{p_t}{(1/2) (p_t + p_u)}
  = \frac{1}{8} (u - t)^2 A''(0) \pm C(A) |u - t|^3,
\end{equation}
where $C(A)$ is a constant depending on the log partition function
$A(\cdot)$, and the expansion is uniform for $u, t$ in a neighborhood of
0.

Finally, we recall that $t = \delta v^T x$ and $u = \delta w^T x$, and
as $|v^T x| \le \ltwo{v} \ltwo{x}$, we have the lower bound
\begin{equation*}
  \inf_p \left\{\dkl{p_{\delta v}(\cdot \mid x)}{p}
  + \dkl{p_{\delta w}(\cdot \mid x)}{p} \right\}
  \ge \frac{1}{8} A''(0) \delta^2 (x^T (w - v))^2
  - C(A) \delta^3 d^{3/2} \max\{\ltwo{w}, \ltwo{v}\}^3.
\end{equation*}
Substituting this into our lower bound~\eqref{eqn:separation-lower-bound}
and using that $\E[XX^T] = I_d$ by construction then gives the lemma.

\subsubsection{Proof of Lemma~\ref{lemma:kl-bound}}
\label{sec:proof-kl-bound}

Without loss of generality, assume that $\ltwo{v} \ge \ltwo{w}$.  We have
\begin{align*}
  \dkl{P_{\delta v}}{P_{\delta w}}
  & = \E\left[\dkl{p_{\delta v}(\cdot \mid X)}{p_{\delta w}(\cdot \mid X)}
    \right].
\end{align*}
Fix $x$ temporarily, and consider the inner KL-divergence term.
As in the proof of Lemma~\ref{lemma:separation}, we use the shorthands
$t = \delta v^T x$, $u = \delta w^T x$, $p_t = p_{\delta v}(\cdot \mid x)$
and $p_u = p_{\delta w}(\cdot \mid x)$, noting that
$|t| \le \delta \sqrt{d} \ltwo{v}$ and similarly for $u$. Then
writing $\E_t$ for expectation under $p_t$, we have
\begin{align*}
  \dkl{p_t}{p_u}
  & = \E_t\left[Y(t - u)\right] + A(u) - A(t)
  = A(u) - A(t) - A'(t)(u - t)
  = \half A''(\wt{u}) (u - t)^2,
\end{align*}
where $\wt{u} \in [u, t]$.  As $A$ is $\mc{C}^\infty$ near $0$, we obtain
that for a constant $C(A)$ depending only on $A$ that
\begin{equation*}
  \dkl{p_t}{p_u} \le \half A''(0) (u - t)^2 + C(A) |u - t|^3,
\end{equation*}
valid for all $u, t \in [-\delta \sqrt{d} \ltwo{v}, \delta \sqrt{d}
  \ltwo{v}]$. We we obtain
\begin{equation*}
  \dkl{P_{\delta v}}{P_{\delta w}}
  \le \frac{\delta^2}{2} A''(0) \E[(X^T (v - w))^2]
  + C(A) \delta^3 \E[|X^T (v - w)|^3],
\end{equation*}
and using $\E[(X^T (v - w))^2] = \ltwo{v - w}^2$ and
$\E[|X^T v|^3] \lesssim \ltwo{v}^3$ for $X
\sim \uniform(\{-1, 1\}^d)$ gives the lemma.


%% file: sections/appendix-bvm.tex

\subsection{Proof of Theorem \ref{thm:bvm}}
\label{appendix:bvm}

Recall $\empmin = \argmin_{\theta \in \Theta} \risk_n(\theta)$ and
Definition~\ref{def:bvmclass}. For shorthand, we use the standard empirical
process notation that $Pf = \E_{P}[f]$ and $P_{n}f = \frac{1}{n} \sum_{i =
  1}^n f(X_{i},Y_{i})$. Let
$\delta_n > 0$ be any sequence satisfying
\begin{equation*}
  \frac{\log \log n}{n} \ll \delta_n^2
  \ll \frac{1}{\sqrt{n}}.
\end{equation*}
We will define a ``good event,'' which is roughly that the empirical risk
$\risk_n$ approximates the true risk $\risk_P$ well and a local quadratic
approximation to both is accurate, and perform our analysis (essentially)
conditional on this good event.  To that end, let $\lambdamin =
\lambdamin(\nabla^2 \risk_P(\theta\opt)$ and $\lambdamax =
\lambdamax(\nabla^2 \risk_P(\theta\opt))$ be the minimum and maximum
eigenvalues of $\nabla^2 \risk_P(\theta\opt)$.  Recall that $\epsilon_1 > 0$
is the radius of the ball on which $\nabla^2 \loss(p_\theta(\cdot \mid x),
y)$ is $\liphess(x, y)$ Lipschitz (Def.~\ref{def:bvmclass}),
and for an $\epsilon > 0$ to be determined
\begin{equation}
  \label{eqn:goodevent}
  \begin{split}
    \goodevent & \defeq \\
    & \left\{ P_{n} \liphess \leq 2 P \liphess,
    ~ \frac{\lambda_{\min}}{2} I \preceq \nabla^2 \risk_n(\theta)
    \preceq 2 \lambda_{\max} I ~ \mbox{for~}
    \ltwo{\theta - \theta\opt} \le \epsilon,
    ~
    \ltwos{\empmin - \theta\opt} \le \delta_n \right\}.
  \end{split}
\end{equation}

We prove the theorem in a series of lemmas. The first shows
that $\goodevent$ occurs eventually, and the remainder we will
demonstrate hold on the event.
\begin{lemma}
  For all sufficiently small $\epsilon > 0$,
  $\goodevent$ happens eventually. That is,
  there is a (random) $N$, finite with
  probability 1, such that $\goodevent$
  occurs for all $n \ge N$.
\end{lemma}
\begin{proof}
  By the strong law of large numbers, we have $P_n \liphess \cas P
  \liphess$, so that $P_n\liphess \le 2P\liphess$ eventually, while
  Definition~\ref{def:bvmclass} implies that $\opnorms{\nabla^2
    \risk_n(\theta) - \nabla^2 \risk_n(\theta\opt)} \le 2 P \liphess \epsilon$
  for all $\ltwo{\theta - \theta\opt} \le \epsilon_1$ on the same
  event. Whenever $\epsilon$ is small enough that
  $P \liphess \epsilon \le \frac{\lambdamax}{2}$ and $P
  \liphess \epsilon \le \frac{\lambdamin}{4}$, we then obtain that
  $\frac{\lambdamin}{2} I \preceq \nabla^2 \risk_n(\theta)
  \preceq 2 \lambdamax I$ by choosing
  \begin{equation*}
    \epsilon \le \min\left\{\epsilon_1,
    \frac{\lambdamin}{4 P \liphess} \right\}.
  \end{equation*}
  Finally, we argue that $\ltwos{\empmin - \theta\opt} \le \delta_n$
  eventually. A standard argument~\cite[Thm.~5.7]{VanDerVaart98}
  and the Glivenko Cantelli theorem, which implies
  $\sup_{\theta \in \Theta} |\risk_n(\theta) - \risk_P(\theta)| \cas 0$
  by the compactness of $\Theta$,
  gives the consistency $\empmin \cas \theta\opt$.
  As $\theta\opt \in \interior \Theta$, Taylor's theorem implies that
  \begin{equation*}
    0 = \nabla \risk_n (\empmin) = \nabla \risk_n (\theta\opt)
    + (\nabla^{2} \risk_n (\theta\opt)+ E_{n}(\empmin, \theta\opt))(
    \empmin - \theta\opt),
  \end{equation*}
  where $E_n$ is an error matrix that Definition~\ref{def:bvmclass} implies
  satisfies
  \begin{equation*}
    \opnorm{E_{n}} \leq \frac{1}{n} \sum_{i=1}^n \liphess(X_i, Y_i)
    \ltwos{\empmin - \theta\opt}.
  \end{equation*}
  Thus $\opnorm{E_n} \cas 0$, and as $\nabla^2 \risk_n(\theta\opt)
  \cas \nabla^2 \risk(\theta\opt)$, we have
  $\empmin - \theta\opt = -(\nabla^2 \risk(\theta\opt) + E_n')^{-1} \nabla
  \risk_n(\theta\opt)$, where $E_n' \cas 0$ is an error matrix.
  By the a.s.\ convergence $E_n' \to 0$
  and law of the iterated logarithm,
  \begin{align*}
    \lefteqn{\limsup_n \sqrt{\frac{n}{\log \log n}} \ltwo{(\nabla^2 \risk(\theta\opt)
      + E_n')^{-1} \nabla \risk_n(\theta\opt)}} \\
    & \qquad \le \opnorm{\nabla^2 \risk(\theta\opt)^{-1}} \limsup_n \sqrt{\frac{n}{\log
        \log n}} \ltwo{\nabla \risk_n(\theta\opt)}
    < \infty
  \end{align*}
  with probability 1. In particular, whenever $\delta_n^2 \gg \frac{\log \log
    n}{n}$, we have $\ltwos{\empmin - \theta\opt} \le \delta_n$ eventually.
\end{proof}

An immediate consequence of the identifiability
condition~\eqref{cond:identifiability} in Definition~\ref{def:bvmclass}
and Taylor's theorem is the following lemma.
\begin{lemma}
  \label{lemma:risk-eigenbounds}
  For all large enough $n$, on event $\goodevent$ we have
  \begin{equation*}
    \risk_n(\theta) \le \risk_n(\what{\theta}_n) + 2 \lambdamax
    \ltwos{\theta - \what{\theta}_n}^2
    ~~ \mbox{for~all~} \ltwos{\theta - \what{\theta}_n} \le \delta_n
  \end{equation*}
  and
  \begin{equation*}
    \risk_n(\theta) \ge \risk_n(\what{\theta}_n) + \frac{1}{4} \lambdamin
    \delta_n^2
    ~~ \mbox{for~all~} \theta \in \Theta ~ \mbox{s.t.~}
    \ltwos{\theta - \what{\theta}_n} \ge \delta_n.
  \end{equation*}
\end{lemma}

\newcommand{\posterior}{\pi}  
\newcommand{\term}{T}  

Finally, we show that on $\goodevent$ we have
\begin{equation*}
  \tvnorm{\vovkmu - \normal\Big(\what{\theta}_n, \frac{1}{n} \nabla^2
    \risk_n(\what{\theta}_n)^{-1}\Big)} \to 0.
\end{equation*}
For shorthand, let $\posterior_n$ be the probability distribution
$\normal(\what{\theta}_n, \frac{1}{n} \nabla^2 \risk_n(\what{\theta}_n)^{-1})$.
We split the variation distance into two terms. Let
$B_n = \delta_n \ball_2^d$ be an $\ell_2$ ball
of radius $\delta_n$. Then
\begin{equation}
  \label{eqn:terms-to-bound}
  2 \tvnorm{\vovkmu - \posterior_n}
  = \underbrace{\int_{\what{\theta}_n + B_n}
    |d\vovkmu - d\posterior_n|}_{
    \eqdef \term_1}
  + \underbrace{\int_{\Theta \setminus \{\empmin + B_n\}}
    |d\vovkmu - d\posterior_n|}_{
    \eqdef \term_2}
  + \underbrace{\posterior_n(\Theta^c)}_{\eqdef \term_3}.
\end{equation}
We bound each of the terms $\term_i$ in turn.  For the second term, we
compute bounds on the densities themselves. Let
$\theta \in \Theta \setminus \{\empmin + B_n\}$.
Then for any $c > 0$ small
enough that $\frac{\lambdamin}{4} - 2 c^2\lambdamax \eqdef K > 0$,
\begin{align*}
  \frac{d}{d\theta} \vovkmu(\theta)
  = \frac{\exp(-n \risk_{n}(\theta))}{
    \int_{\Theta} \exp(-n \risk_n(\theta') )d \theta'}
  & \leq \frac{\exp(-n \risk_{n}(\theta))}{
    \int_{\what{\theta}_n + c B_n}  \exp(-n \risk_n(\theta') )d \theta'} \\
  & \stackrel{(i)}{\leq}
  \frac{\exp(-\frac{\lambdamin}{4} n \delta_{n}^2)}{
    \exp(-2 \lambdamax c^{2} \delta_{n}^{2}) \vol(c B_n)}
  = \exp\left(- n K \delta_{n}^{2} + d \log \frac{1}{c \delta_{n}}
  - c_d \right),
\end{align*}
where inequality~$(i)$ follows from Lemma~\ref{lemma:risk-eigenbounds}
and $c_d = \log \vol(\ball_2^d)$ is the log volume of the $\ell_2$-ball.
A completely analogous calculation gives
\begin{align*}
  \frac{d}{d \theta} \posterior_n(\theta)
  & =
  \frac{\exp(-\frac{n}{2} (\theta - \empmin)^{T} \emphess (\theta - \empmin))}{
    \int \exp(-\frac{n}{2} (\theta' - \empmin)^{T}
    \emphess (\theta' - \empmin)) d \theta ' } \\
  & \leq  \frac{\exp(-\frac{\lambdamin}{4} n \delta_{n}^{2})}{
    \exp(-2 \lambdamax c^{2} \delta_{n}^2) \vol(c B_n)}
  = \exp\left(- n K \delta_n^2 + d \log \frac{1}{c \delta_n}
  - c_d\right),
\end{align*}
where the inequality uses the definition~\eqref{eqn:goodevent} of
$\goodevent$.  In particular, setting
the constant $c = \frac{1}{4} \sqrt{\frac{\lambdamin}{\lambdamax}}$,
the term $K = \frac{1}{8} \lambdamin$ and
we may bound term $\term_2$ in
expression~\eqref{eqn:terms-to-bound} by
\begin{equation*}
  \term_2 \le 2 \vol(\Theta)
  \exp\left(-n K \delta_n^2 + \frac{d}{2} \log \frac{16 \lambdamax}{
    \lambdamin \delta_n^2} - c_d \right) \to 0,
\end{equation*}
as $\delta_n \gg \frac{1}{\sqrt{n}}$ and $\delta_n \to 0$.

Let us turn to term $\term_1$ in expression~\eqref{eqn:terms-to-bound}.
For sets $A \subset \R^d$ we define the normalizing constants
\begin{equation*}
  Z_{A,n}^{\normal}
  \defeq \int_{A} \exp\left(-\frac{n}{2} (\theta - \empmin)^{T} \emphess
  (\theta - \empmin) \right) d \theta
  ~~\mbox{and}~~
  Z_{A,n}^{\mathsf{Vovk}} \defeq
  \int_{A} \exp\left(-n( \risk_{n}(\theta) - \risk_n(\empmin))\right) d \theta.
\end{equation*}
Changing notation slightly to let $B_n = \empmin + \delta_n \ball_2^d$,
Lemma~\ref{lemma:risk-eigenbounds}
implies the inequalies
\begin{equation*}
  \max\left\{Z^{\mathsf{Vovk}}_{\Theta \setminus B_n, n},
  Z^{\normal}_{\Theta \setminus B_n, n}
  \right\} \leq  \vol(\Theta)
  \exp\left(-\frac{\lambdamin}{4} n \delta_{n}^{2}\right)
\end{equation*}
and
\begin{equation*}
  \min\left\{Z^{\normal}_{B_n,n},
  Z^{\mathsf{Vovk}}_{B_n, n}\right\}
  \ge \exp(-2 n \lambdamax c^2 \delta_n^2) \vol(c \delta_n \ball_2^d),
\end{equation*}
valid for any $c \le 1$. Thus, the ratio
\begin{align*}
  \rho_n
  \defeq
  \max\left\{\frac{Z_{\Theta \setminus B_n, n}^{\mathsf{Vovk}}}{Z_{B_n, n}^\normal},
  \frac{Z_{\Theta \setminus B_n, n}^{\normal}}{Z_{B_n, n}^{\mathsf{Vovk}}}\right\}
  & \le
  \frac{\vol(\Theta) \exp(-\frac{\lambdamin }{4} n \delta_{n}^2)}{
    \exp(-2 n \lambdamax c^2 \delta_{n}^2) \vol(c \delta_n \ball_2^d)} \\
  & \le \vol(\Theta) \exp\left(-n \delta_n^2
  \left(\frac{\lambdamin}{4} - 2 c^2 \lambdamax\right) +
  d \log \frac{1}{c \delta_n} - c_d\right),
\end{align*}
where as before $c_d = \log \vol(\ball_2^d)$,
so that for all small $c > 0$ we have $\rho_n \to 0$ as $n \to \infty$
on the event $\goodevent$.
We may then bound the normalizing constant ratio
by
\begin{equation}
  \frac{Z_{B_n,n}^{\mathsf{Vovk}}}{Z_{B_n,n}^{\normal}}
  + \rho_n
  \ge \frac{Z_{B_n,n}^{\mathsf{Vovk}} + Z_{\Theta \setminus B_n, n}^{\mathsf{Vovk}}}{
    Z_{B_n,n}^{\normal}}
  \ge \frac{Z_{\Theta,n}^{\mathsf{Vovk}}}{Z_{\Theta,n}^{\normal}}
  \ge \frac{Z_{B_n,n}^{\mathsf{Vovk}}}{Z_{B_n, n}^{\normal}
    + Z_{\Theta \setminus B_n, n}^{\normal}}
  \ge \left(\frac{Z_{B_n, n}^{\normal}}{Z_{B_n,n}^{\mathsf{Vovk}}}
  + \rho_n \right)^{-1}.
  \label{eqn:normalizing-ratios}
\end{equation}
Performing a Taylor expansion, on $\goodevent$, for any
$\theta \in B_n$ the Lipschitz continuity of $\nabla^2 \risk_n(\theta)$ implies
\begin{equation*}
  \risk_n(\theta) = \risk_n(\empmin)
  + \half (\theta - \empmin)^T \emphess (\theta - \empmin)
  \pm P \liphess \cdot \delta_n^3.
\end{equation*}
Using this $O(\delta_n^3)$ remainder term, we then immediately
obtain the ratio bounds
\begin{equation*}
  \frac{Z_{B_n,n}^{\mathsf{Vovk}}}{Z_{B_n,n}^{\normal}}
  =
  \frac{\int_{B_n}
    \exp\left(-n (\risk_{n}(\theta) - \risk_n(\empmin))\right) d \theta}{
    \int_{B_n} \exp( - \frac{n}{2} (\theta- \empmin)^{T} \emphess
    (\theta - \empmin)) d \theta}
  \in \exp\left(\pm P \liphess \cdot \delta_n^3\right).
\end{equation*}
Substituting this containment in the inequalities~\eqref{eqn:normalizing-ratios},
we find that for all large enough $n$, on the event
$\goodevent$ in Eq.~\eqref{eqn:goodevent}, we have the bounds
\begin{equation}
  \label{eqn:ratio}
  \exp(-P \liphess \delta_n^3) - O(1) \rho_n
  \le \frac{Z_{\Theta,n}^{\mathsf{Vovk}}}{Z_{\Theta,n}^{\normal}}
  \le \exp(P \liphess \cdot \delta_n^3) + O(1) \rho_n.
\end{equation}
Finally, we return to computing the densities in the term $\term_1$
in Eq.~\eqref{eqn:terms-to-bound}.
Let $Z_n^\normal = Z_{\R^d,n}^\normal$,
where an argument similar to those above shows that
$Z_n^\normal / Z_{\Theta,n}^\normal \to 1$ as $n \to \infty$.
Defining the remainder
$\mbox{rem}_n(\theta) = \risk_n(\theta) - \risk_n(\empmin)
- \half (\theta - \empmin)^T \nabla^2 \risk_n(\empmin) (\theta - \empmin)$
and using that $\ltwos{\mbox{rem}_n(\theta)} \le P \liphess \cdot \delta_n^3$
for any $\theta \in B_n$ as above,
the inequalities~\eqref{eqn:ratio} imply
\begin{align*}
  \lefteqn{\left|d\vovkmu(\theta) - d \posterior_n(\theta) \right| / d\theta
    = \exp\left(-\frac{n}{2} (\theta - \empmin)^{T} \emphess (\theta - \empmin)
    \right)
    \left| \frac{\exp(- n r_{n}(\theta))}{Z_{\Theta,n}^{\mathsf{Vovk}}}
    -  \frac{1}{Z_n^{\normal}} \right|} \\
  & \le \frac{ \exp(-\frac{n}{2} (\theta - \empmin)^{T} \emphess
    (\theta - \empmin) )}{Z_{n}^{\normal}}
  \left[\left| \exp(-n P \liphess \cdot \delta_n^3)
    \frac{Z_{\Theta, n}^{\normal}}{Z_{n}^{\mathsf{Vovk}}} - 1\right|
    + \left|\frac{1}{Z_n^{\normal}} - \frac{1}{Z_{\Theta,n}^\normal}\right|
    \right]
\end{align*}
Integrating over $B_n$ and invoking inequality~\eqref{eqn:ratio}
then implies
\begin{align*}
  \term_1 = \int_{B_n} |d\vovkmu - d\posterior_n|
  & \le \frac{\int_{B_n}  \exp(-\frac{n}{2} (\theta - \empmin)^{T} \emphess
    (\theta - \empmin) ) }{Z_{n}^\normal}
  \cdot o(1) \to 0.
\end{align*}

Lastly, we note that the final term $\term_3$ in the variation
distance~\eqref{eqn:terms-to-bound} satisfies $\posterior_n(\Theta^c) \to 0$
as $n \to \infty$ as on event $\goodevent$, there is eventually a ball of
some (fixed) radius $\epsilon > 0$ such that $\what{\theta}_n + \epsilon
\ball_2^d \subset \Theta$, and $\nabla^2 \risk_n(\empmin) \succeq
(\lambdamin/2) I$. For Standard normal concentration results then
immediately imply that $\posterior_n(\Theta^c) \le
\posterior_n(\{\what{\theta}_n + \epsilon \ball_2^d\}) \to 0$, as the
variance of $\theta \sim \posterior_n$ satisfies
$\E_{\posterior_n}[\ltwo{\theta - \E[\theta]}^2] \le C / n$ for some
problem-dependent $C$.
We conclude that each of $\term_1, \term_2, \term_3 \to 0$ in
the variation distance~\eqref{eqn:terms-to-bound}.

%% file: sections/appendix-bvmcor.tex

\subsection{Proof of Corollary \ref{cor:bvm}}
\label{appendix:corbvm}

We again use the event $\goodevent$ in Eq.~\eqref{eqn:goodevent} in the
proof of Theorem~\ref{thm:bvm} and $\frac{\log \log n}{n} \ll \delta_n^2 \ll
\frac{1}{\sqrt{n}}$ as well.  Let $p_n = \vovkP[n]$ and $\mu_n = \vovkmu[n]$
for shorthand, and let $p_{\empmin}$ the the point model. Let $B_n = \empmin
+ n^{-1/4} \ball_2^d$ be a ball of radius $n^{-1/4}$ around $\empmin$, where
for all large enough $n$, on $\goodevent$ we have $B_n \subset B \subset
\Theta$, where we recall that $B$ is the neighborhood of $\theta\opt$ in
Assumption~\ref{assumption:extra}. Then for the base measure $\nu$ on
$\mc{Y}$, we expand
\begin{align*}
  2 \tvnorm{p_n(\cdot \mid x) - p_{\empmin}(\cdot \mid x)}
  & = \int \left|\int_\Theta \left(p_\theta(y \mid x) - p_{\empmin}(y \mid x)
  \right) d\mu_n(\theta) \right| d\nu(y) \\
  & \le \mu_n(\Theta \setminus B_n)
  + \int \left|\int_{B_n} \left(p_\theta(y \mid x) - p_{\empmin}(y \mid x)
  \right) d\mu_n(\theta) \right| d\nu(y).
\end{align*}
By Theorem~\ref{thm:bvm}, we have
$\mu_n(\Theta \setminus B_n) \to 0$ on $\goodevent$.
Now, let $\score_\theta = \log p_\theta$
for shorthand, and also define the shorthands
$\dot{p}_\theta = \nabla_\theta p_\theta$
and $\dot{\score}_\theta = \nabla_\theta \score_\theta =
\frac{\dot{p}_\theta}{p_\theta}$.
The Lipschitz condition on $\log p_\theta$ in Assumption~\ref{assumption:extra}
guarantees that (for large $n$) on the set $B_n$ we have
$|\frac{\dot{p}_\theta(y \mid x)}{p_\theta(y \mid x)}| \le \lipP(x, y)$
for $\theta \in B_n$. Writing
\begin{equation*}
  p_\theta(y \mid x) - p_{\empmin}(y \mid x)
  = \int_0^1 \dot{p}_{t \theta + (1 - t) \empmin}(y \mid x)^T
  (\theta - \empmin) dt
  = \int_0^1 \dot{\score}_{t \theta + (1 - t) \empmin}(y \mid x)^T (\theta - \empmin)
  p_{t \theta + (1 - t) \empmin}(y \mid x) dt,
\end{equation*}
we have
\begin{equation*}
  |p_\theta(y \mid x) - p_{\empmin}(y \mid x)|
  \le \lipP(x, y) \ltwos{\theta - \empmin}
  \int_0^1 p_{t \theta + (1 - t) \empmin}(y \mid x) dt.
\end{equation*}
Thus
\begin{align*}
  \lefteqn{\int_{\mc{Y}}
    \left|\int_{B_n} \left(p_\theta(y \mid x) - p_{\empmin}(y \mid x)
    \right) d\mu_n(\theta) \right| d\nu(y)} \\
  & \le \int_{\mc{Y}} \int_{B_n} \lipP(x, y)
  \ltwos{\theta - \empmin} \int_0^1 p_{t \theta + (1 - t) \empmin}(y \mid x)
  dt d \mu_n(\theta) d\nu(y) \\
  & = \int_0^1 \int_{B_n} \ltwos{\theta - \empmin}
  \left[\int_{\mc{Y}} \lipP(x, y) p_{t \theta + (1 - t) \empmin}(y \mid x)
    d\nu(y)
    \right] d\mu_n(\theta) dt
  \le \lipP(x) n^{-1/4}
\end{align*}
on $\goodevent$, and we have the desired convergence.